# Theory-guided Hard Constraint Projection (HCP): a Knowledge-based Data-driven Scientific Machine Learning Method

Yuntian Chen[a], Dou Huang[b], Dongxiao Zhang[c,*], Junsheng Zeng[a], Nanzhe Wang[d], Haoran Zhang[e,f,b], and Jinyue Yan[e]

[a]Intelligent Energy Laboratory, Frontier Research Center, Peng Cheng Laboratory, Shenzhen, 518000, P. R. China
[b]Center for Spatial Information Science, The University of Tokyo, Chiba, 277-8568, Japan
[c]School of Environmental Science and Engineering, Southern University of Science and Technology, Shenzhen, 518055, P. R. China
[d]BIC-ESAT, ERE, and SKLTCS, College of Engineering, Peking University, Beijing, 100871, P. R. China
[e]Future Energy Center, Malardalen University, Vasteras, 721 23, Sweden
[f]LocationMind Inc., Tokyo 101-0032, Japan
*Corresponding author. E-mail address: zhangdx@sustech.edu.cn

**Abstract:** Machine learning models have been successfully used in many scientific and engineering fields. However, it remains difficult for a model to simultaneously utilize domain knowledge and experimental observation data. The application of knowledge-based symbolic artificial intelligence (AI) represented by an expert system is limited by the expressive ability of the model, and data-driven connectionism AI represented by neural networks is prone to produce predictions that might violate physical principles. In order to fully integrate domain knowledge with observations and make full use of the strong fitting ability of neural networks, this study proposes theory-guided hard constraint projection (HCP). This deep learning model converts physical constraints, such as governing equations, into a form that is easy to handle through discretization and then implements hard constraint optimization through projection in a patch. Based on rigorous mathematical proofs, theory-guided HCP can ensure that model predictions strictly conform to physical mechanisms in the constraint patch. The training process of theory-guided HCP only needs a small amount of labeled data (sparse observation), and it can supervise the model by combining the coordinates (label-free data) with domain knowledge. The performance of the theory-guided HCP is verified by experiments based on a published heterogeneous subsurface flow problem. The experiments show that theory-guided HCP requires fewer data and achieves higher prediction accuracy and stronger robustness to noisy observations than the fully connected neural networks and soft constraint models. Besides, due to the application of domain knowledge, theory-guided HCP has the ability to extrapolate and can accurately predict points outside the range of the training dataset.

# 1. Introduction
## 1.1 Symbolic AI, connectionism AI, and knowledge-based data-driven model

In the past few centuries, many fields in science and engineering have accumulated rich domain knowledge and prior information. Knowledge in these fields has played an important role in theoretical studies and practical applications. Since the Dartmouth Conference in 1956, the development of artificial intelligence (AI) has gone through two stages: symbolic AI, represented by expert systems; and connectionism AI, represented by data-driven models [1]. Symbolic AI utilizes domain knowledge through expert systems [2, 3] and knowledge engineering [4] to develop some early artificial intelligence systems, such as DENDRAL, for organic chemical structure analysis [5], MYCIN for diagnosis of blood infectious diseases [6], and Deep Blue that defeated the world chess champion Kasparov [7]. Although these symbol-based methods are able to use domain knowledge, they can only solve problems in a complete information and structured environment [1]. However, since many situations in the real world cannot be described simply by rules, symbolic AI has limitations in practice. In addition, symbolic AI lacks a solid mathematical foundation and can only rely on the usage of simple logic. Therefore, the learning and fitting capability of these models might be insufficient to describing many complex problems in practice, and can only solve specific problems.

A common method of describing complex real-world scenarios is the data-driven model with strong fitting capabilities (i.e., connectionism AI). Perceptron is a kind of typical connectionism AI model [8], which is characterized by its strong fitting ability, but it relies heavily on training data. Typical achievements include certain models, such as convolutional neural networks [9], Hopfield networks [10], long short-term memory [11], and graph convolutional networks [12]. Connectionism AI is the current mainstream method of deep learning, and has been successfully applied in numerous fields [13, 14]. Connectionism AI contains many data-driven machine learning models that depend heavily on training data. In addition, obtaining data is always expensive and time-consuming in actual scientific and engineering scenarios. The observations are also always incomplete (i.e., the training data cannot cover all application scenarios) and contain inevitable noise in practice. Essentially, data-driven models do not use domain knowledge accumulated over decades and the model does not have "common sense" or understanding of the real world (physical mechanisms). This lack of understanding of domain knowledge makes it easy for commonly used neural networks to generate unreasonable or unrealistic predictions. For example, Dong et al. [15] found that the convolutional neural network does not truly detect semantic objects (e.g., the model does not learn the concept of a bird when identifying birds from images). Therefore, for some artificial adversarial samples, such as specially designed images of beer bottles, neural networks will mistake them for birds although they have a totally different visual appearance. Similarly, in the problem of subsurface flow, previous studies have shown that the conventional data-driven neural network exhibits poor performance on observations with noise, which is inevitable in practice [16] and it cannot accurately predict the flow field in the future, since these time steps are not included in the training data. Therefore, the neural network might produce predictions that do not conform to the physical mechanism at the boundaries of the flow field and the areas where the value space changes drastically.

In order to solve the problem of the lack of common sense in the currently commonly used data-driven models, the academic community has attempted to integrate domain knowledge and prior information into the machine learning model to achieve a dual-driven model of knowledge and data. For example, the U.S. Department of Energy proposed scientific machine learning in 2019 [17], and the concept of third-generation artificial intelligence was advanced in 2020 [1]. The essence of machine learning models is to extract and use information as efficiently as possible for inference and prediction. Symbolic AI (e.g., expert systems) tends to focus only on domain knowledge, while connectionism AI (e.g., artificial neural networks) tends to focus only on data. The strategy to fundamentally improve the performance of a machine learning model lies in the enhancement of its input information, i.e., richer data and more essential prior information or domain knowledge. Only by allowing the model to use domain knowledge and data information at the same time, can it be ensured that the model not only exerts strong fitting ability of the data-driven model, but also possesses the stability of the expert system. Moreover, the prediction results of the knowledge and data dual-driven model are more likely to be consistent with physical mechanisms.

**1.2 Soft constraint model**

Converting domain knowledge into the constraints of the model optimization process is a feasible method for using the domain knowledge. In the field of constrained optimization, researchers attempt to solve the constrained problem by modifying the architecture of the neural networks [18, 19]. However, the non-convex nature of neural networks makes direct optimization of the constraints difficult. To solve this problem, Pathak el at. proposes a two-step training strategy to decouple the constraints from the network output in the field of computer vision [20]. At each iteration, they compute a latent probability distribution as the closest point in the constrained region. Then, they update the network parameters to follow the latent probability distribution as closely as possible. The constraints in this model are very simple, such as the existence and the size of different labels in the images. Furthermore, a theory-guided framework was proposed from a deep learning perspective and achieved good performance in the field of fluid dynamics [21, 22] and hydrology [16, 23, 24]. The essence of this model framework is to use first principle models and empirical models as the reference and basis for model prediction. The term "theory-guided" refers to the model framework that combines data-driven methods and model-driven methods. Theory-guided data science (TGDS) was proposed by Karpatne et al. in 2017 [25]. Karpatne et al. also advanced the physics-guided neural network (PGNN) in 2017 [26]. A similar method, called the physics informed neural network (PINN), was developed by Raissi et al. in 2019 [21], in which nonlinear partial differential equations are utilized as regularization terms in the loss function. These terms essentially reflect prior information formed by domain knowledge, aiming to characterize the data understanding obtained in advance. In addition, Beucler et al. proposed a method of enforcing nonlinear constraints in neural networks by changing them to linear constraints in architecture or the loss function [27]. Wang et al. developed the theory-guided neural network (TgNN) based on TGDS and PINN, in which the governing equation of underground seepage is taken as domain knowledge to guide neural network predictions [16]. Under the theory-guided framework, researchers further proposed models, including TgDLF [28] and TgLSTM [29], to solve the problem of fusing knowledge and data in practical scenarios. Compared with pure data-driven models that can only interpolate within the range of training data, theory-guided models can perform extrapolation based on physical mechanisms and predict data outside of the range of training data, which improves the application scope of the model.

The aforementioned theory-guided and physics-informed models ensure that the prediction results are close to the physical mechanism by embedding the governing equation in the loss function as regularization terms. These methods are essentially soft constraints, which count the weighted average of the degree of consistency between the model predictions and the physical mechanism. The regularization terms can only weakly enforce the constraints and guarantee that the prediction results conform to the physical constraints in the average sense, and they cannot ensure that the predictions do not violate the physical constraint at each point (i.e., the limitation of penalty method [30]). In fact, there are sometimes even physically unreasonable predictions that deviate significantly from the governing equation in practice.

**1.3 Hard constraint model**

From an optimization perspective, hard constraints are more efficient methods than soft constraints, in general. Hard constraints can ensure that a given governing equation is strictly satisfied in a certain area or an entire domain. Theoretically, the application of hard constraint enables the model to require fewer data, achieve higher prediction accuracy, and stronger robustness to noisy observations. Current studies on hard constraints in deep learning are still preliminary. Physics constrained learning (PCL) is an attempt to introduce hard constraints [30]. It embeds constraint by solving PDE, which can ensure that the predictions satisfy the governing equation. Its challenge lies in how to ensure that the gradient can be effectively propagated in the PDE solver. Therefore, PCL requires complex mathematical derivation for specific problems in practice. PCL solves the governing equation on the basis of the chain rule, and the neural network is only used as a substitute model for the complex functions in the governing equation. Therefore, PCL is more similar to a numerical simulation method, not a deep learning model. Mohan et al. advanced a method to embed hard constraints via convolutional neural network (CNN) kernels in 2020 [31]. In this method, the neural network predicts unlabeled intermediate variables and then obtains the output value through a physics-embedded decoder (which includes many CNN kernels corresponding to differential operations [32]). Essentially, the physics-embedded decoder performs a full-field iteration on the entire physical field. This method is enlightening, but still necessitates a complete solution to the entire field, which is similar to an iterative step in simulation. In order to avoid solving the entire physical field, a feasible way is to project the flow field locally to the hyperplane determined by constraints. In terms of constructing projections based on domain knowledge, Chen and Zhang developed a method of training neural networks with the assistance of domain knowledge through projection in 2020 [33], but this method can only embed the linear mapping relationship between different variables into the neural network for constructing labels in the training data. As a consequence, the proposed projection method cannot handle complex partial differential equations.

In this study, we attempt to incorporate the governing equations as hard constraints into the neural networks through projection under the theory-guided framework. Specifically, the governing equation is discretized and the predictions in a small area in the flow field are extracted as a constraint patch. Subsequently, the governing equation is decomposed into a constraint matrix reflecting the physical mechanism and the prediction matrix composed of neural network outputs on each constraint patch. Finally, a hyperplane conforming to the governing equations is constructed based on the constraint matrix and the prediction matrix is projected to the hyperplane to obtain predictions that satisfy the physical constraints. Since the final outputs are on the hyperplane, the projected results strictly conform to the governing equation (i.e., hard constraint) on the constraint patch. The proposed model is trained end-to-end.

This paper comprises four sections. In section 2, we propose the hard constraint projection (HCP) to transform domain knowledge into hard constraints and construct a theory-guided HCP to generate predictions that meet physical constraints. In section 3, the heterogeneous subsurface flow problem is taken as an example to analyze the difference between hard constraints and soft constraints through experiments and evaluate the performance of theory-guided HCP under different quantities and qualities of training data. Finally, this study is summarized and discussed in section 4.

## 2. Methodology

In order to intuitively compare the overall architecture of the conventional data-driven neural network, soft constraint based neural network, and hard constraint based neural network, Figure 1 shows the illustrations of the three models. The input variables of the model are three-dimensional space-time coordinates (x, y, t), and only some of the coordinates have observations as the labels (i.e., partial and sparse observation), which are represented as solid-line cubes in the figure.

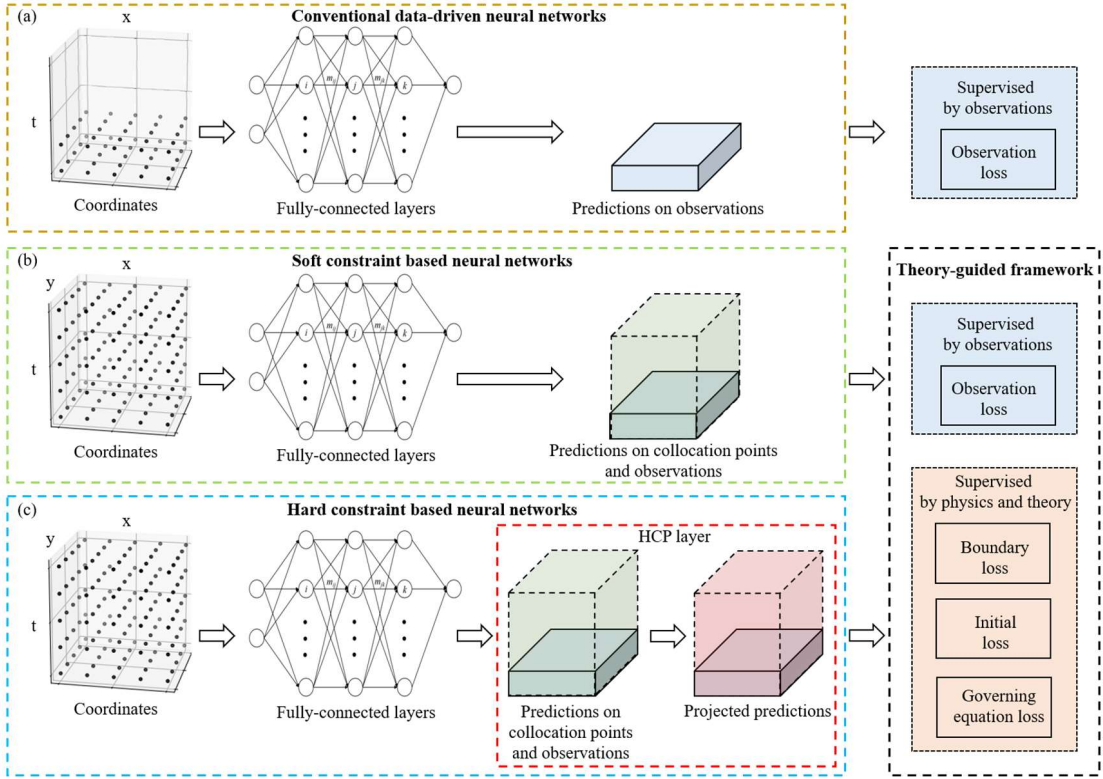

**Fig. 1.** Illustration of overall architecture of (a) conventional data-driven neural network; (b) soft constraint based neural network; and (c) hard constraint based neural network. The theory-guided framework and the HCP layer is shown in the black dashed box and red dashed box, respectively.

The conventional data-driven neural network in Figure 1a can fit the mapping relationship between input coordinates and corresponding observations and train the model through observation loss (i.e., supervised by observations). The soft-constrained neural network in Figure 1b and the hard-constrained neural network in Figure 1c use the theory-guided framework (black dashed box), which facilitates the model to comprehensively optimize model parameters based on observations and theories. Therefore, the output data in Figure 1b and Figure 1c contains not only the observations represented by the solid-line cubes, but also the predictions on the collocation points (dotted cubes) that are used to calculate the boundary loss, initial loss, and governing equation loss. The theory-guided framework can not only use the observation error to train the model, but also use theory to guide the learning process of the network. This framework is introduced in detail in section 2.1. In addition, the hard constraint neural network in Figure 1c has a specially designed HCP layer, where the hard constraint projection (HCP) can be regarded as an activation function based on domain knowledge. The HCP layer can ensure that the network prediction results are in line with the given constraints. The details of the HCP layer are introduced in section 2.2.

**2.1 Theory-guided framework**

In order to introduce domain knowledge, especially the physical mechanism that has been refined into concise governing equations, into the model, the theory-guided neural network has recently been developed to bridge data-driven and model-driven approaches [16]. These networks adopt a theory-guided framework whose essence is to use first principle models and empirical models as the guide and reference for model predictions. In the theory-guided framework, the governing equations are converted into constraints and introduced into the loss function of the model as penalty terms (or regularization terms). These penalty terms essentially reflect prior information formed by domain knowledge, aiming to characterize the data understanding obtained in advance.

The illustration of the theory-guided framework is shown in Figure 2. The blue box on the left denotes the observation loss, describing the difference between the predictions and the observations. The conventional data-driven model builds a loss function based on this loss to optimize the model. In addition to the observation loss, the theory-guided framework also includes governing equation loss, boundary loss, and initial loss (shown in the orange box) based on domain knowledge (such as the governing equation in the gray box). Finally, the theory-guided loss function is obtained by linear combination of all of the penalty terms mentioned above.

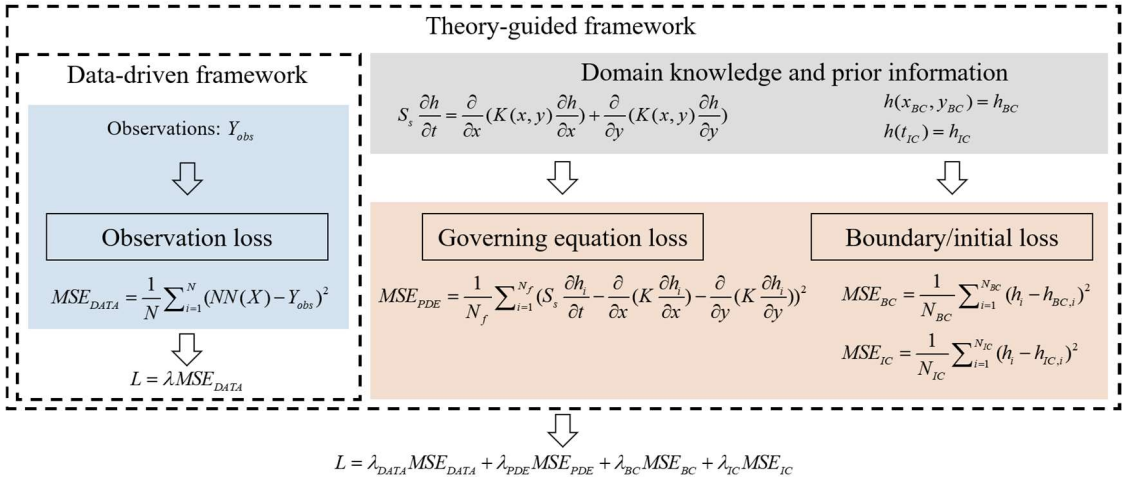

**Fig. 2.** Illustration of theory-guided framework.

Specifically, a neural network can be expressed as Eq. 1. If we know a certain physical relationship between some input variables and the target (e.g., conservation of energy or conservation of mass), and this relationship can be described in mathematical form, such as Eq. 2, we can then introduce this prior knowledge as a constraint to the model:

$$Y = NN(X) = f(X) \tag{1}$$

Subject to $Y = g(X'),$ where $X' \subseteq X$ \hfill (2)

where $X$ and $Y$ are the independent variables and the dependent variables, respectively; $NN(\bullet)$ denotes the neural network; $f(\bullet)$ represents the mapping relationship fitted by the neural network; $g(\bullet)$ is the mathematical representation of domain knowledge, such as governing equations; and $X'$ denotes a subset of the set of input variables $X$.

It should be noted that Eq. 2 is often in the form of ordinary differential equation (ODE) or partial differential equation (PDE) in practice. Therefore, it is challenging, or even impossible, to predict the target $Y$ based on a specific $X'$ when the $Y$ values of the surrounding area are unknown and a time-consuming simulation is always required. In addition, since $X'$ is a subset of $X$, Eq. 2 only describes part of the system (e.g., only describes conservation in a partial domain) and the information contained in Eq. 2 is usually insufficient to directly predict the target $Y$.

Although Eq. 2 is not a sufficient condition for predicting the targets, it can be utilized to improve the performance of the model as a necessary condition. In order to utilize the governing equation as a constraint, we rewrite Eq. 2 and obtain Eq. 3 by moving all non-zero terms to the left side. If the neural network predictions strictly conform to the constraint (Eq. 2), then Eq. 3 holds; otherwise, the left side of Eq. 3 is not 0. Therefore, it is straightforward to take the value on the left side of Eq. 3 as an index to describe the degree of deviation between the neural network predictions and the constraint, i.e., $g'(X',Y)$ can be used as a penalty term in the loss function of the neural network in the theory-guided framework:

$$g'(X',Y) = 0 \tag{3}$$

In this study, the problem of subsurface flow in the field of fluid mechanics is taken as an example to better demonstrate the theory-guided framework. Eq. 4 is the governing equation for this problem and it can be taken as domain knowledge to guide neural network predictions:

$$S_s \frac{\partial h}{\partial t} = \frac{\partial}{\partial x}(K(x,y)\frac{\partial h}{\partial x}) + \frac{\partial}{\partial y}(K(x,y)\frac{\partial h}{\partial y}) \tag{4}$$

where $S_s$ denotes the specific storage; $K(x, y)$ denotes the hydraulic conductivity field; and $h$ denotes the hydraulic head.

The penalty terms of governing equation ($MSE_{PDE}$), boundary condition ($MSE_{BC}$), and initial condition ($MSE_{IC}$) are defined as Eq. 5, which are added to the loss function to avoid final predictions from violating physical mechanisms:

$$MSE_{PDE} = \frac{1}{N_f} \sum_{i=1}^{N_f} (g'(X',Y) - 0)^2 = \frac{1}{N_f} \sum_{i=1}^{N_f} (S_s \frac{\partial h_i}{\partial t} - \frac{\partial}{\partial x}(K \frac{\partial h_i}{\partial x}) - \frac{\partial}{\partial y}(K \frac{\partial h_i}{\partial y}))^2$$

$$MSE_{BC} = \frac{1}{N_{BC}} \sum_{i=1}^{N_{BC}} (h_i - h_{BC,i})^2 \tag{5}$$

$$MSE_{IC} = \frac{1}{N_{IC}} \sum_{i=1}^{N_{IC}} (h_i - h_{IC,i})^2$$

where $N_f$, $N_{BC}$, and $N_{IC}$ denote the number of collocation points, boundary points, and initial points, respectively; $h_{BC}$ and $h_{IC}$ are the hydraulic head of the boundary condition and initial condition, respectively; and $h_i$ represents the i$^{th}$ prediction of the neural network.

Finally, by introducing the penalty terms in Eq. 5 into the loss function, the theory-guided loss can be obtained as follows:

$$L = \lambda_{DATA} MSE_{DATA} + \lambda_{PDE} MSE_{PDE} + \lambda_{BC} MSE_{BC} + \lambda_{IC} MSE_{IC} \tag{6}$$

where $\lambda$ are hyperparameters, which control the weights of each term in the loss function; and $MSE_{DATA}$ denotes the prediction error of the neural network compared to the observations.

It should be mentioned that the "theory" in the theory-guided framework contains multiple sources of information, including not only governing equations, but also expert experience and engineering constraints. The theory refers to the information summarized based on prior information and domain knowledge. The theory-guided framework is flexible and it can be compatible with highly complex theories and mechanisms, as long as they can be expressed in mathematical form.

**2.2 Hard constraint modeling process**

From an optimization point of view, the theory-guided framework introduced in section 2.1 takes domain knowledge and prior information as soft constraints. However, models based on soft constraints may still produce physically-inconsistent results since the regularization terms in the loss function can only guarantee that the prediction results do not seriously violate the constraints, in the average sense. In general, hard constraints are more desired in optimization since they can ensure that a given governing equation is strictly satisfied in a certain area, which is termed a constraint patch.

The construction of hard constraint models is always more challenging than soft constraint models. In order to embed domain knowledge into neural networks as hard constraints, this study proposes a three-step projection method, including equation discretization, matrix decomposition, and projection. The process of equation discretization is similar to finite element analysis in computational fluid dynamics, which can transform the governing equation into a discrete form that is easier to handle. Matrix decomposition divides the discretized equation into a prediction matrix composed of neural network predictions and a constraint matrix that represents the physical mechanism, so as to optimize the prediction matrix under the premise of satisfying the physical constraints. Projection is done to map the values in the prediction matrix to the hyperplane, conforming to the constraint matrix to obtain physically reasonable prediction results (i.e., conforming to the physical mechanism).

### 2.2.1 Equation discretization and matrix decomposition

In the process of discretization, the partial differential equation is transformed into a differential structure based on the idea of finite difference. For example, the governing equation of subsurface flow (Eq. 4) is discretized based on the second-order center difference scheme along the x and y dimensions and the first-order backward Euler scheme along the t dimension. Finally, the discretized equation of the subsurface flow is obtained as Eq. 7:

$$0 = \frac{S_S}{\Delta t} h^{t-\Delta t} + (-\frac{S_S}{\Delta t} + \frac{-(K_{x+\Delta x/2} + K_{x-\Delta x/2})}{\Delta x^2} + \frac{-(K_{y+\Delta y/2} + K_{y-\Delta y/2})}{\Delta y^2}) h^t \\ + \frac{K_{x-\Delta x/2}}{\Delta x^2} h^t_{x-\Delta x} + \frac{K_{x+\Delta x/2}}{\Delta x^2} h^t_{x+\Delta x} + \frac{K_{y-\Delta y/2}}{\Delta y^2} h^t_{y-\Delta y} + \frac{K_{y+\Delta y/2}}{\Delta y^2} h^t_{y+\Delta y} \quad (7)$$

where $x$, $y$, and $t$ denote the coordinates of a point in space and time, respectively; and $\Delta x$, $\Delta y$, and $\Delta t$ denote the difference interval along the $x$, $y$, and $t$ directions, respectively.

In order to describe the influence of physical constraints on the local flow field (or other physical fields), we introduce the concept of the constraint patch. The constraint patch is the surrounding area of a given point where the distance between adjacent points is equal to the difference interval ($\Delta x$, $\Delta y$, and $\Delta t$). The discretized equation Eq. 7 describes the physical mechanisms that the model should obey. Only when the predictions of all points in the constraint patch meet the relationship described in Eq. 7, does the constraint patch meet the constraint of the governing equation Eq. 4. It should be mentioned that the direct predictions of the neural network in the constraint patch may not satisfy the discretized equation Eq. 7, since the theory-guided framework are essentially soft constraints. We need to adjust the predictions in the constraint patch through projection, so that the adjusted results are not only the closest results to the original predictions, but also are in line with the constraint that corresponds to the governing equation Eq. 4.

The matrix decomposition is performed to effectively project the predictions. Specifically, the predictions of the neural network in the constraint patch are extracted as the prediction matrix (Eq. 8). The remaining part of the discretized equation can be constructed into a constraint matrix (Eq. 9), which is determined based on physical parameters (e.g., hydraulic conductivity) and the coordinates in the constraint patch. The constraint matrix is essentially the hyperplane constrained by the discretized equation in the variable space, reflecting the relationship determined by the physical mechanism at different positions in the constraint patch.

$$H = \begin{bmatrix} h^{t-\Delta t}, & h^t, & h^t_{x-\Delta x}, & h^t_{x+\Delta x}, & h^t_{y-\Delta y}, & h^t_{y+\Delta y} \end{bmatrix}^T \quad (8)$$

$$A = \begin{bmatrix} \frac{S_S}{\Delta t}, & a_2, & \frac{K_{x-\Delta x/2}}{\Delta x^2}, & \frac{K_{x+\Delta x/2}}{\Delta x^2}, & \frac{K_{y-\Delta y/2}}{\Delta y^2}, & \frac{K_{y+\Delta y/2}}{\Delta y^2} \end{bmatrix}$$

$$a_2 = -\frac{S_S}{\Delta t} + \frac{-(K_{x+\Delta x/2} + K_{x-\Delta x/2})}{\Delta x^2} + \frac{-(K_{y+\Delta y/2} + K_{y-\Delta y/2})}{\Delta y^2} \quad (9)$$

where $H$ is the prediction matrix; and $A$ is the constraint matrix that denotes the physical constraints.

## 2.2.2 Hard constraint projection (HCP)

The essence of hard constraint projection is to project the prediction matrix obtained by the neural network to the hyperplane determined by the constraint matrix in the high dimensional feature vector space. The projected result is an adjusted matrix closest to the original prediction matrix and satisfies the given constraints.

From a mathematical perspective, the above task can be expressed as: For any $A \in \mathbb{R}^{m,n}$, where $m < n$, and $rank(A) = m$. Given any $w \in \mathbb{R}^n$, find the closest point $w^*$ subject to $Aw^* = b$.

This problem can be solved by projection, the proof of which is provided in Appendix A. The finally obtained projection matrix is defined as $P = (I - A^T(AA^T)^{-1}A)$, which is the key of hard constraint projection. In the experiment of section 3, the projection matrix can ensure that the difference between the projected predictions and the governing equation in the constraint patch is at the level of $10^{-10}$, which indicates that the projected predictions strictly conform to the differential format of governing equation in the constraint patch. According to Eq. 8 and Eq. 9, Eq. 7 can be rewritten as $AH = 0$. Therefore, the prediction matrix $H$ can be projected to a hyperplane that conforms to a certain physical mechanism (Eq. 7) based on Theorem 2 in Appendix A. The adjusted prediction matrix is obtained as follows:

$$H\_ = PH = (I - A^T(AA^T)^{-1}A)H, \text{ where } AH\_ = 0 \tag{10}$$

where $H\_$ denotes the projected prediction matrix; and $P$ denotes the projection matrix.

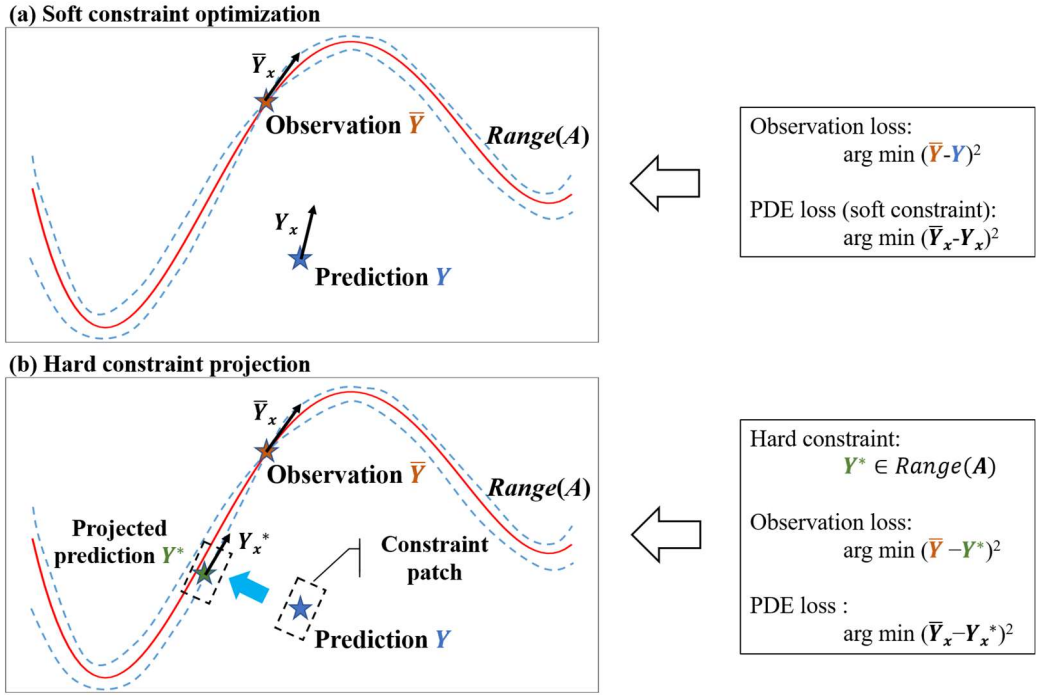

**Fig. 3.** Illustration of (a) soft constraint optimization; and (b) hard constraint projection (HCP).

In order to intuitively show the projection process and compare the difference between the HCP

and soft constraint model (TgNN), an illustration of the optimization process of the two models is provided in Figure 3. The red curve represents the hyperplane determined by the physical constraint, i.e., $Range(A)$, where $A$ is the constraint matrix. The blue star represents the original neural network prediction, which deviates from the physical constraint (red curve). The brown star represents the true observation. Since the observations meet the physical constraints, the brown star is on the red curve. The green star indicates the projection result of prediction on $Range(A)$.

It should be noted that the projection process is operated on the constraint patch (black dashed box) around the prediction. Since the constraint patch can only represent the local information of the physics constraint, there are multiple solutions (blue dashed lines). These blue lines denote the local constraint hyperplanes determined by the physical constraint in the constraint patch. They are close to the red line, which globally satisfies the constraint, and all local and global constraint hyperplanes intersect at the brown star (i.e., the observation is a general solution). HCP can convert the predictions in the constraint patch into projections (green star) that meet physical constraints in the patch. In an ideal situation, if the information in the constraint patch approximately contains the global constraint information, then the projections obtained based on the local constraint hyperplane also satisfies the global constraint hyperplane (i.e., the blue dashed lines and the red line coincide and the green star is on the red line).

The black arrows in the figure represent the gradients at the corresponding positions. The gradient at observation corresponds to $\bar{Y}_x$, the gradient at prediction corresponds to $Y_x$, and the projection corresponds to $Y_x^*$. The TgNN in Figure 3a is a typical soft constraint model. Its optimization process is to minimize the distance between the predictions and the observations ($\bar{Y}$-$Y$) and the difference between the gradients ($\bar{Y}_x$-$Y_x$). Figure 3b shows the optimization process of HCP. First, the predictions in the constraint patch are projected to the hyperplane, determined by the physical constraint (red curve). Then, the distance between the projected predictions (green star) and the observations (brown star) ($\bar{Y}$-$Y^*$) and the difference between their gradients ($\bar{Y}_x$-$Y_x^*$) are minimized. Since the projected predictions (green star) are on $Range(A)$, the predictions of HCP strictly conform to the physical constraints in the constraint patch.

The detailed calculation process of hard constraint projection is presented in Figure 4. First, we randomly take the target point from the neural network predictions and extract the constraint patch around the target point. When the target point is at the boundary of the physical field, the constraint patch is constructed by setting ghost cells outside of the boundary. We then adjust the prediction matrix through the projection matrix of Eq. 10 to obtain the prediction matrix in line with the physical mechanism. Finally, the results of all constraint patches are spliced together to obtain physically reasonable prediction results in the entire field.

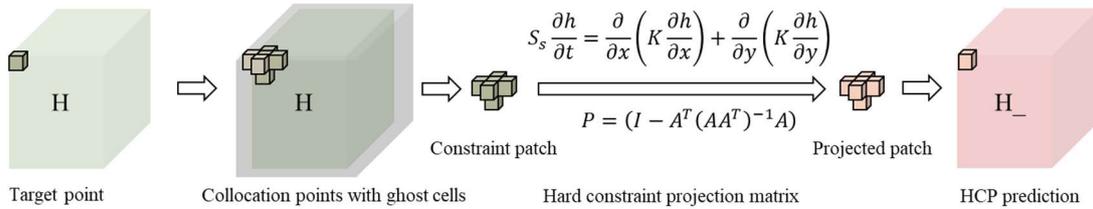

**Fig. 4.** Work flow of hard constraint projection.

In actual applications, boundary conditions are the key to solving the governing equations, but direct projection through Eq. 10 cannot guarantee that the projected results meet the boundary conditions. Consequently, a special projection strategy is required for the boundaries. Specifically, we introduce the concept of the ghost cell in computational fluid dynamics (CFD) to deal with the boundary conditions. Ghost cells are virtual grid cells outside of the boundary, similar to the padding method in computer vision. By specifying the relationship between the ghost cells and the boundary cells, we can constrain the boundary cells to obtain a solution that satisfies the boundary conditions. Appendix B shows the method of using ghost cells to handle the commonly-used constant pressure boundary and no-flow boundary. The projections of other boundary conditions can also be achieved using a similar method.

**2.3 Theory-guided hard constraint projection**

In order to make full use of domain knowledge and prior information to train neural networks, this study proposes theory-guided HCP. On the one hand, this method can implement hard constraints on the predictions through projection and ensure that the model predictions will not violate the physical mechanism. On the other hand, the flexibility of the theory-guided framework is conducive to fusing various information from the model training process, which is significant for improving the model accuracy. The model architecture of theory-guided HCP is illustrated as a flow chart in Figure 5. The theory-guided HCP is similar to conventional neural networks. It consists of two parts: the feedforward inference process with hard constraints; and the backward process with backpropagation of theory-guided loss.

In the feedforward process, the input of the neural network is a set of coordinates ($x$, $y$, $t$), corresponding to the location of the training data in the physical field. The predictions of the neural network consist of two parts, corresponding to samples with observations (solid-line cube) and collocation points without observations (dotted cubes). The network predictions of this step do not necessarily satisfy the physical constraints, which calls for the hard constraint projection introduced in section 2.2. In order to obtain physically reasonable predictions, the governing equation is discretized and converted into a projection matrix and the original predictions of the neural network are adjusted in the constraint patch through hard constraint projection. Intuitively, the projected results constitute the closest solution to the original predictions in the value space that satisfies the governing equation since the theory-guided HCP utilizes the projection to adjust the original prediction results. In the backward process, the theory-guided HCP calculates the training loss under the theory-guided framework, which comprehensively considers the observations and domain knowledge, such as boundary condition, initial condition, and governing equation.

It should be mentioned that when the theory-guided HCP utilizes the governing equation to constrain the model, it only requires the coordinates (inputs) of the data at the boundary and the randomly selected constraint patches. The training process of theory-guided HCP only needs a small amount of labeled data and it can supervise the neural network by combining label-free data with domain knowledge (i.e., collocation points in Section 3.31). Furthermore, compared to conventional deep learning algorithms, the observation data used by theory-guided HCP do not need to cover the entire field. In other words, theory-guided HCP has the ability to extrapolate and is capable of predicting points outside the range of the training dataset. For instance, in the experiment introduced in section 3, there are a total of 50 time steps, and only the first 18 time steps have observation data, and the remaining 32 time steps are label-free data (i.e., sparse observation or partial observation). This problem is challenging for conventional deep learning algorithms because these algorithms require training data and prediction data to have similar data distributions and they cannot guarantee the performance for extrapolation. However, the theory-guided HCP possesses the ability to extrapolate due to the introduction of domain knowledge through hard constraints and the theory-guided framework, which is verified in section 3. This feature indicates the broader applicability of the model in real applications.

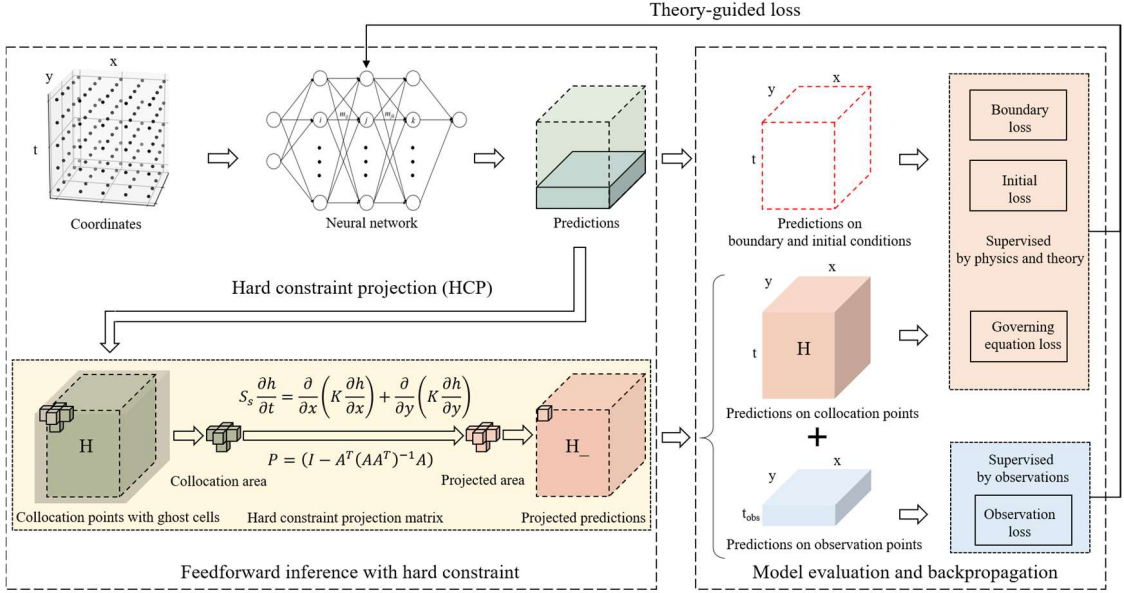

**Fig. 5.** Flow chart of theory-guided hard constraint projection (HCP).

In addition, compared to solving the physical field through numerical simulation, theory-guided HCP has the advantage of data-driven algorithms that require less calculation in prediction. For example, to predict the result of $t$ time step at location ($x$, $y$), it is not necessary to calculate from the 1$^{st}$ time step to the $t^{th}$ time step, nor does it need to start from the boundary. Theory-guided HCP only needs to input the coordinate (t, x, y) as independent variables into the model to obtain the results. Appendix C provides a comparative experiment on the computational time of theory-guided HCP and simulation to solve a physical field.

In summary, theory-guided HCP takes advantage of the theory-guided framework and introduces physical constraints through the loss function, which can effectively use domain knowledge to assist model training. Meanwhile, theory-guided HCP converts the soft constraints in the existing models into hard constraints through discretization and projection, which can ensure that the results obtained in the constraint patch strictly conform to the governing equation. The conformity of the model predictions with the physical mechanism assists to enhance the reliability and credibility of the model.

**2.4 Intuitive understanding of theory-guided HCP**

The difference between hard constraints and soft constraints lies in whether or not an improvement suggestion is provided to obtain a solution that meets the constraint. In soft constraints, the algorithm evaluates the degree of deviation between the predictions and the constraints (e.g., a governing equation) and uses this deviation as the loss in the iterative optimization process. Therefore, soft constraints are essentially a qualitative measure of whether the model predictions are closer to, or farther from, the physical constraint. However, the predictions in the constraint patch are transformed into values that meet physical constraints by projection in the theory-guided HCP. Consequently, the hard constraints not only evaluate the degree of deviation between the predictions and the physical constraints, but also quantitatively improve the prediction results.

Intuitively, if the optimization process is regarded as climbing down a mountain (Figure 6), it is assumed that there is a river (red curve) flowing from the top of the mountain to the valley and climbing along the river is assumed to be the fastest path. The lowest point (brown) denotes the global minimum (observation), the green point denotes the initial prediction (starting point for optimization), and the river (red curve) represents the constraint hyperplane determined by the constraint patch. In the data-driven method, the algorithm takes the direction with the fastest gradient decrease at each position as the optimization direction, so that it goes along the white triangles and finally converges to a local minimum. In soft constraint methods, the altitude of the current position (i.e., the deviation of the prediction from the observation) and the distance from the river (i.e., the deviation from the constraints) are measured and combined as a loss value in the optimization process. The optimization process of soft constraints is done to balance the need to decrease the altitude with the need to be closer to the river, which cannot guarantee that the result of each iteration step meets the physical constraint in the collocation point or constraint patch. The blue diamonds in Figure 6 illustrate the optimization process of the soft constraint methods. For hard constraint methods, the closest position along the river (yellow star) to each step is found through projection and then the loss is measured in the optimization process. As a consequence, the projected predictions (yellow stars) can be guaranteed to meet the physical constraints (red curve) in the constraint patch (i.e., every foothold must be along the river).

Compared with soft constraints, hard constraints adjust the optimization direction of each step through projection, so that every prediction must be in line with the governing equation in the constraint patch (i.e., the yellow stars are always on the red curve). This is an intuitive understanding of theory-guided HCP.

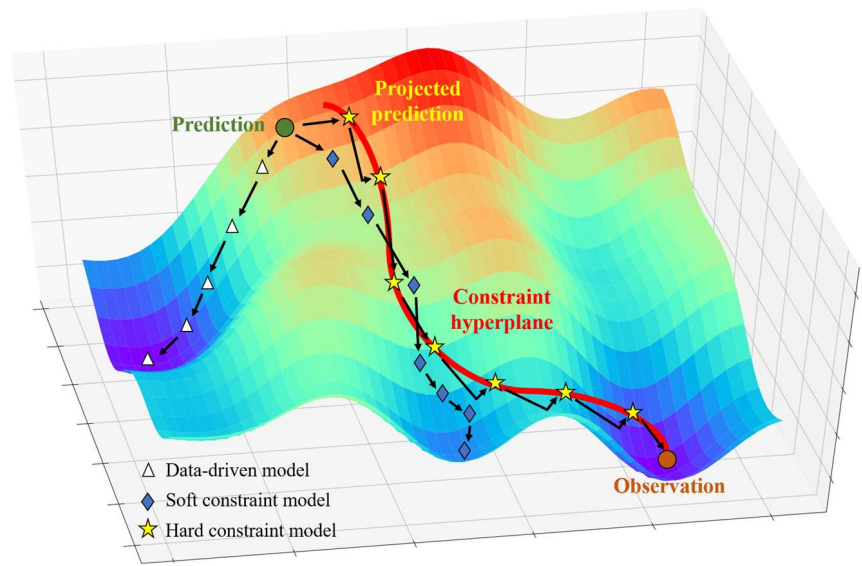

**Fig. 6.** Illustration of optimization process (climbing down a mountain) via data-driven model (white triangle), soft constraint model (blue diamond), and hard constraint model (yellow star). The red curve denotes the constraint hyperplane (river). The green point denotes the initial prediction (starting point for optimization).

## 3. Experiment
### 3.1 Heterogeneous subsurface flow problem

In this section, in order to thoroughly evaluate the performance of the proposed theory-guided HCP, the problem of subsurface flow in saturated porous medium is taken as a case study and multiple experiments are carried out. The subsurface flow problem in these experiments should satisfy the governing equation given in Eq. (4). The hydraulic conductivity field $K(x, y)$ in the governing equation is set to be heterogeneous to better simulate real situations in practice. Considering that the heterogeneous fields are regarded as random fields, the Karhunen–Loeve expansion (KLE) could be utilized to randomly generate the heterogeneous fields following a specific distribution with corresponding covariance [34, 35].

In this study, a two-dimensional transient saturated flow in porous medium is considered. In order to objectively and thoroughly evaluate the performance of theory-guided HCP in subsequent experiments, the heterogeneous field in these experiments uses the same settings as those in a published TgNN study [16]. Specifically, the domain is a square, which is evenly divided into 51×51 grid blocks and the length in both directions is 1020 [$L$], where [$L$] denotes any consistent length unit. The left and right boundaries are set as constant pressure boundaries and the hydraulic head takes values of $H_{x=0}$=202 [$L$] and $H_{x=1020}$=200 [$L$], respectively. Furthermore, the two lateral boundaries are assigned as no-flow boundaries. The grid and the boundaries are illustrated in Figure B1. The specific storage is assumed as a constant, taking a value of $S_S$=0.0001 [$L^{-1}$]. The total simulation time is 10 [$T$], where [$T$] denotes any consistent time unit, with each time step being 0.2 [$T$], resulting in 50 time steps. The initial conditions are $H_{t=0,x=0}$=202 [$L$] and $H_{t=0,x\neq0}$=200 [$L$]. The mean and variance of the log hydraulic conductivity are given as 0 and 1, respectively. In addition, the correlation length of the field is $\eta$=408 [$L$]. The hydraulic conductivity field is parameterized through KLE and 20 terms are retained in the expansion. Therefore, this field is represented by 20 random variables $\xi=\{\xi_1(\tau), \xi_2(\tau),\cdots \xi_{20}(\tau)\}$ in the considered cases. The conductivity field obtained through KLE is shown in Figure 7a, which exhibits strong anisotropy. Finally, MODFLOW [36] software is adopted to perform the simulations to obtain the required dataset, and the data distributions at two time steps are presented in Figure 7b and 7c.

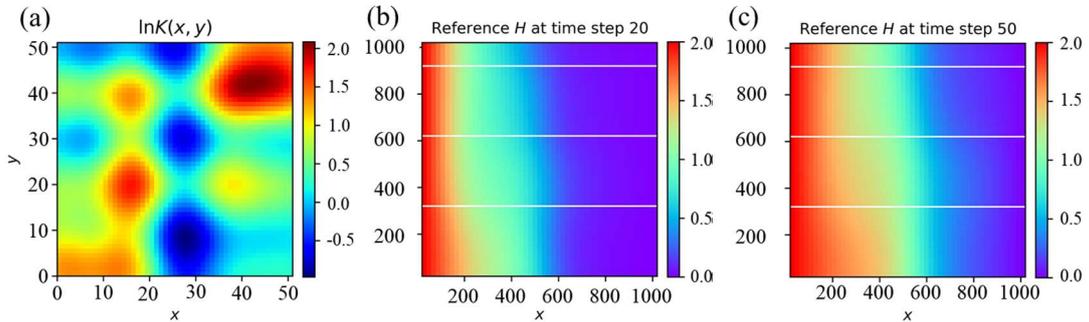

**Fig. 7.** Color maps of conductivity field hydraulic pressure field.

The numerical simulation results contain a total of 50 time steps of the hydraulic head field, of which the data of the first 18 time steps are used as observations in the training dataset and the data of the last 32 time steps are used as the test dataset. We attempt to predict the development of the flow field through domain knowledge and a few observations at early time steps in this study. This constitutes a typical extrapolation problem, and existing studies have proven that directly using a commonly-used artificial neural network (ANN) cannot accurately predict the development of the flow field. The TgNN is the state-of-the-art model for the extrapolation problem in the field of subsurface flow. It utilizes information in the governing equation for the training process through soft constraints, which improves the model's prediction accuracy compared with the ANN. However, a large number of collocation points and boundary points are needed in the training process.

In subsequent experiments, the performance and robustness of different models are evaluated by comparing ANN, TgNN, and theory-guided HCP in the case of using a different number of collocation points, boundary points, and observations, as well as a different noise level and proportion of outliers. The relative L2 loss of each model in the experiments is provided in Appendix C. The relative L2 loss is a dimensionless parameter, which is the ratio of the 2-norm (Euclidean norm) of the difference between the predictions and the observations to the 2-norm of the observations.

**3.2 Difference between hard constraints and soft constraints**

In order to verify the effect of the hard constraints, the decreasing trends of different losses in the first 500 epochs of theory-guided HCP and TgNN are compared, as shown in Figure 8. The theory-guided HCP and TgNN use the same network architecture and are trained based on the same dataset. Figure 8a shows the comprehensive loss value of the models. Figure 8b presents the observation loss, which denotes the deviation from the observation. Figure 8c shows the PDE loss, which denotes the deviation between the predictions and the governing equation (the gradients in the equation are obtained by automatic differentiation of the neural network). Figure 8d presents the condition loss, which is the deviation of the predictions from the boundary conditions and initial conditions. Since the initial value of the neural network is often close to 0, the gradients obtained based on automatic differentiation are almost 0. Since the flow field with all zero values also conforms to the governing equation, the PDE loss is close to 0 in the initial iteration steps in Figure 8c. However, the all-zero flow field does not meet the boundary conditions and initial conditions, so the other losses are quite high in the initial time steps and the PDE loss will increase first and then decrease. Comparing the various loss curves in Figure 8, it can be seen that the loss value of theory-guided HCP decays faster than TgNN, so the method of embedding physical constraints through projection is more effective and instructive in practice.

It should be noted that the adjustment before and after the projection in each step of the theory-guided HCP is often small. Thus, the projection will not have a significant impact on the value of the prediction result in a single step. Figure 9a shows the prediction matrix of 30 randomly extracted constraint patches in one iteration. The abscissa represents different constraint patches and the ordinate corresponds to the six predictions in each constraint patch. Figure 9b presents the adjusted predictions of the constraint patches after projection, and Figure 9c shows the change rate of the prediction matrix before and after the projection. It can be seen that in theory-guided HCP, the prediction matrix can be converted into a projected prediction matrix that meets the physical constraints in a constraint patch by only adjusting the value of less than 1% of the prediction results. Although the difference in value brought by the projection at each iteration step is small, this change can make the theory-guided HCP find the optimization direction more rapidly and efficiently, which often leads to higher prediction accuracy and robustness in practice. In subsequent experiments, we will comprehensively evaluate the performance of theory-guided HCP with hard constraints and TgNN with soft constraints.

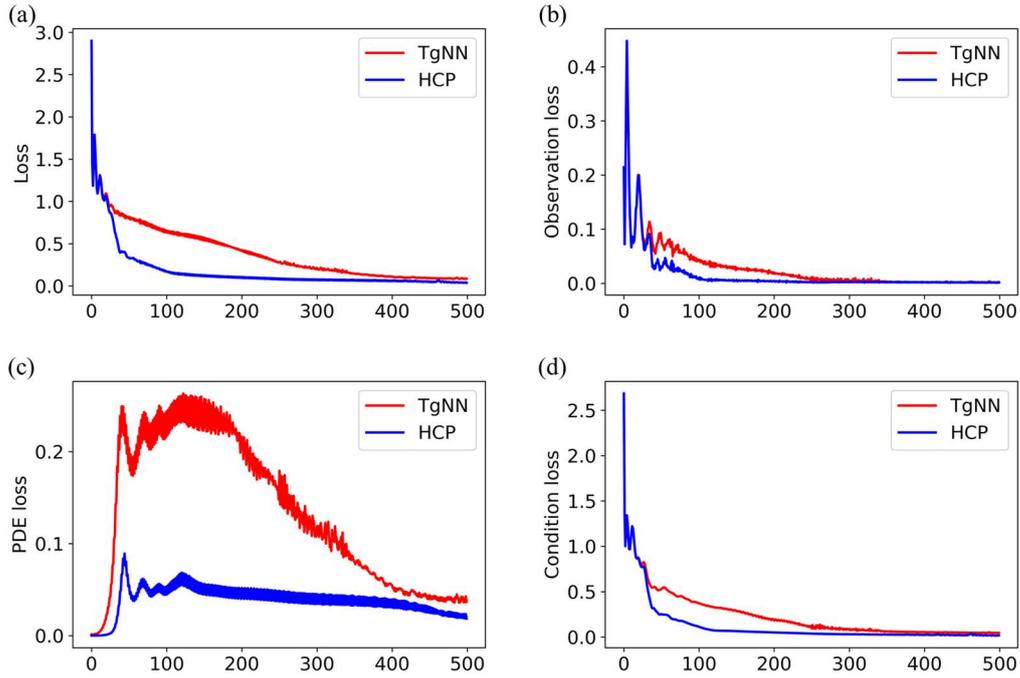

**Fig. 8.** Decreasing trends of losses in the first 500 epochs of TgNN and HCP: (a) comprehensive MSE loss of TgNN (red) and HCP (blue); (b) observation loss; (c) PDE loss; and (d) condition loss.

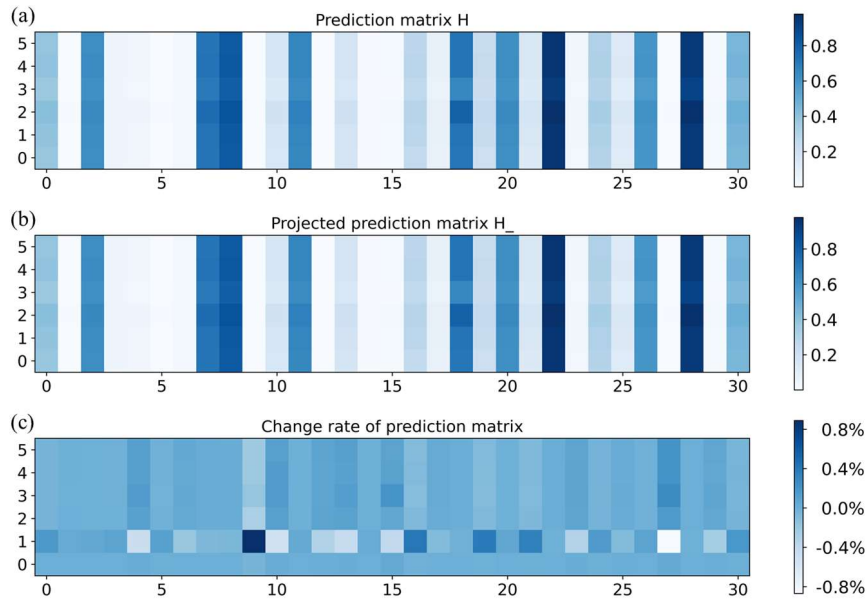

**Fig. 9.** Color map of (a) prediction matrix; (b) projected prediction matrix; and (c) change rate of prediction matrix.

### 3.3 Data demand analysis

Theory-guided HCP is capable of converting governing equations into hard constraints and efficiently utilizes prior information and domain knowledge through projection. Theoretically, it has stronger learning ability and lower data demand than purely data-driven models (e.g., ANN) and soft constraint models (e.g., TgNN). In this section, we analyze the performance of theory-guided HCP, TgNN, and ANN in different numbers of collocation points, boundary points, and observations through comparative experiments. To the best of the authors' knowledge, TgNN is the state-of-the-art model that uses domain knowledge (especially governing equations) for model training and ANN is one of the most widely used deep learning models.

**3.3.1 Predicting the future responses with different numbers of collocation points**

The collocation points are key for the models that utilize information in the governing equation to train the neural network, such as TgNN and PINN. The theory-guided framework takes advantage of the automatic differentiation of neural networks to calculate the derivative of each feature at the collocation points. The difference between the derivatives and the governing equation (Eq. 4) is introduced as a penalty term into the loss function in the training process. Although the penalty term does not depend on observations at the collocation points, the stronger the model's ability to extract information from collocation points, the smaller the number of collocation points required in the training process.

In order to evaluate the effectiveness of the proposed projection method, the performance of theory-guided HCP and TgNN at nine different numbers of allocation points are compared with the same neural network architecture. In order to avoid the influence of randomness on the model evaluation, the experiment was repeated independently and a boxplot was drawn to reflect the data distribution. The experimental results are presented in Figure 10.

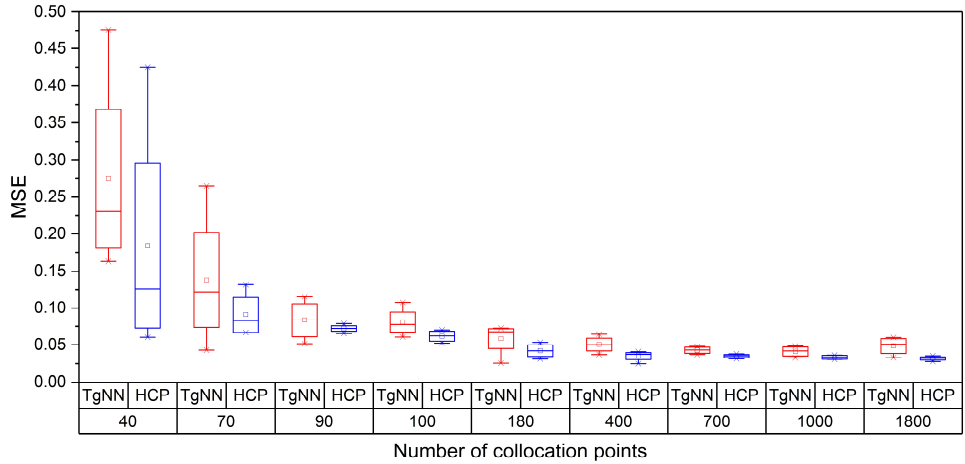

**Fig. 10.** Boxplot of the relative L2 loss of TgNN and HCP with different numbers of collocation points.

It can be seen from the figure that when the number of allocation points is less than 400, the relative L2 loss of the theory-guided HCP prediction is less than that of TgNN, which indicates higher accuracy. In addition, by comparing the height of each box, it is found that the performance of theory-guided HCP in different random experiments has stronger consistency, which indicates that HCP can extract common information from the collocation points and it is less likely to be affected by the unique information of these points. As the number of collocation points increases, the model performance of theory-guided HCP and TgNN gradually converges. Overall, both of the models have similar prediction accuracy and the robustness of theory-guided HCP is superior to that of TgNN.

In order to intuitively show the performance of theory-guided HCP, the prediction results of theory-guided HCP, TgNN, and ANN at the 30[th] time step (Figure 11a) and the 50[th] time step (Figure 11b) with 40 collocation points are compared. The three columns in Figure 11a and Figure 11b correspond to the three horizontal lines with y coordinate of 320, 620, and 920 in the flow field (the total flow field ranges from 0 to 1020). The ordinate in the figure shows the hydraulic head, and the abscissa is the x coordinate in the flow field. It should be mentioned that the data corresponding to the 30[th] and 50[th] time steps have exceeded the range covered by the observation data (the first 18 time steps), i.e., the labeled training data and the data to be predicted have different data distributions. Therefore, there may be patterns that do not appear in the observations and the performance of purely data-driven models (e.g., ANN) is not ideal. Although the performance of all models tends to deteriorate for later time steps, the deviation between the predicted value of theory-guided HCP (red line) and the ground truth (blue line) is always smaller than that of TgNN and ANN. Figure 11c shows the cross plot of the prediction and ground truth of all 50 time steps of theory-guided HCP, TgNN, and ANN. Since Figure 11c contains the data of all time steps, the model performance is more comprehensively evaluated. The predictions of theory-guided HCP are closest to the 45° diagonal, indicating that theory-guided HCP has the highest accuracy. Figure 12 shows the model performance when the number of collocation points is 400. As the number of collocation points increases, the prediction accuracy of theory-guided HCP and TgNN improves, while the results of ANN are not affected.

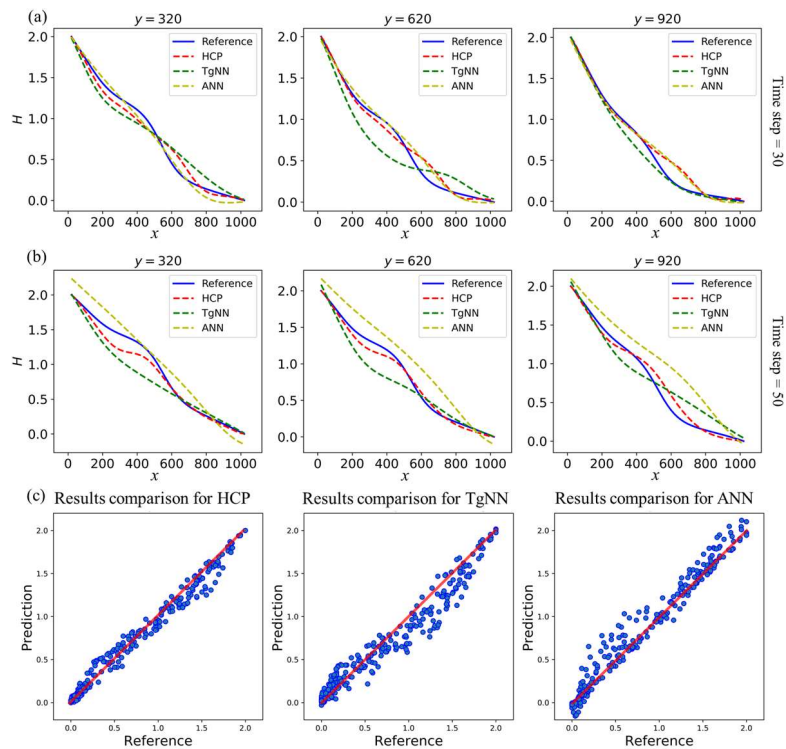

**Fig. 11.** Performance of HCP, TgNN, and ANN with 40 collocation points: (a) prediction results at the 30[th] time step; (b) prediction results at the 50[th] time step; and (c) cross plot of prediction and ground truth of all time steps.

Therefore, theory-guided HCP has higher prediction accuracy and robustness in scenarios with a small number of collocation points, which means that the proposed projection method can more efficiently extract domain knowledge and prior information in the governing equations from a limited number of collocation points.

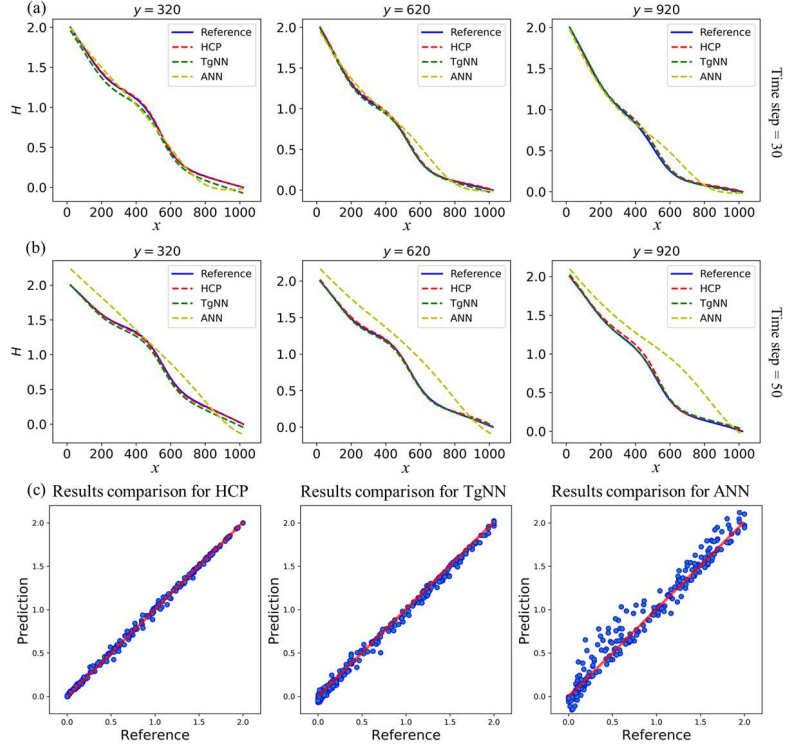

**Fig. 12.** Performance of HCP, TgNN, and ANN with 400 collocation points: (a) prediction results at the 30[th] time step; (b) prediction results at the 50[th] time step; and (c) cross plot of prediction and ground truth of all time steps.

### 3.3.2 Predicting the future response with different numbers of boundary points

In this section, we evaluate the performance of theory-guided HCP and TgNN near the boundary of the field. Figure 13 compares the model performance with different numbers of sampling points on boundaries. The number of sampling points ($N_b$) on each boundary ranges from 1 to 10,000. Considering that each flow field has four boundaries, the total number of boundary points ranges from 4 to 40,000. It is shown in the figure that theory-guided HCP possesses an advantage when the number of boundary points is small (left side of the figure). As the number of boundary points increases, the results of theory-guided HCP and TgNN gradually converge, but even when the number of boundary points reaches 10,000, the hard constraints (HCP) still have higher accuracy and stronger robustness.

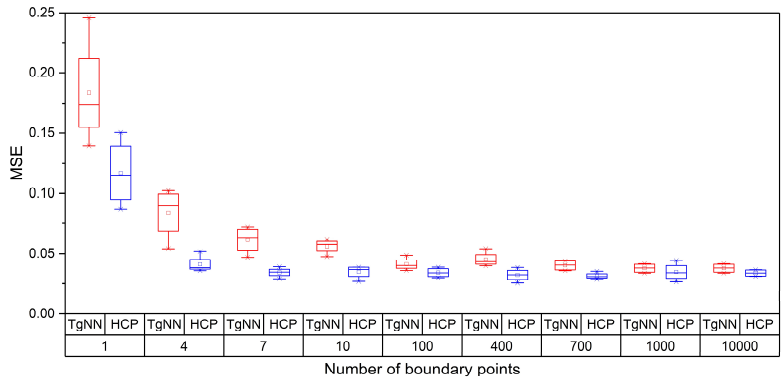

**Fig. 13.** Boxplot of the relative L2 loss of TgNN and HCP with different numbers of boundary points.

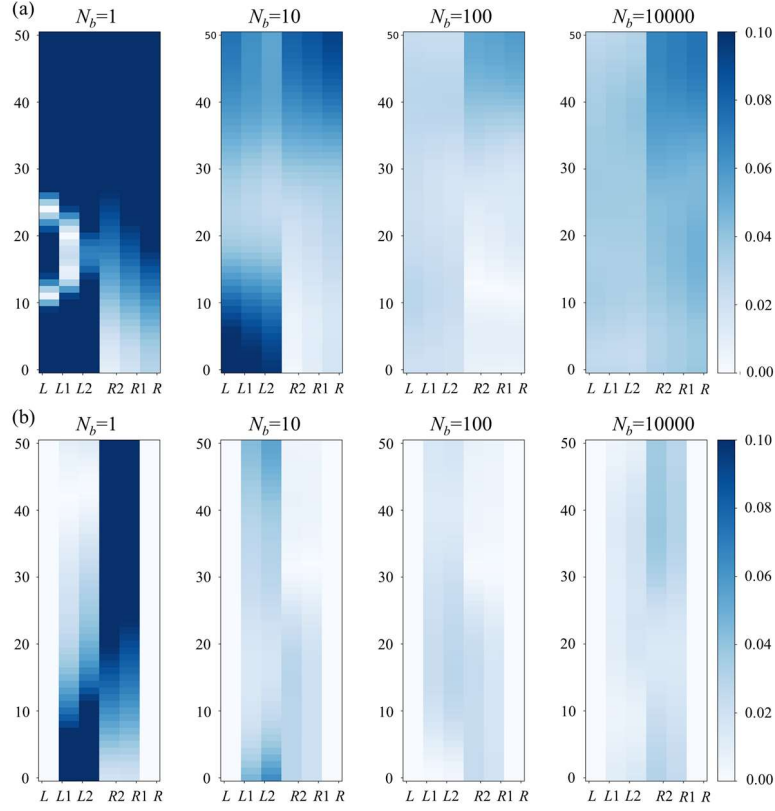

**Fig. 14.** Residual map of (a) TgNN with different numbers of boundary points; and (b) HCP with different numbers of boundary points. L and R represent the left and right boundary grids, respectively. L1 and L2 correspond to the two layers of grids next to the left boundary. R1 and R2 correspond to the two layers next to the right boundary.

In order to more clearly demonstrate the difference between theory-guided HCP and TgNN at the boundaries, we compared the performance of theory-guided HCP and TgNN on the three layers of grids near the boundary of the flow field when the number of boundary points is 1, 10, 100, and 10,000, respectively. Figure 14 shows the difference (residual) between the prediction results and the ground truth. Figure 14a and Figure 14b present the absolute residual maps of TgNN and theory-guided HCP, respectively. All of the residual maps use the same scale of color bar. It can be seen that the color of theory-guided HCP is lighter than the TgNN at the same number of boundary points and that the residuals at the boundaries are small. This experimental result demonstrates that theory-guided HCP achieves better performance at the boundaries, which reflects the advantages of hard constraints in handling boundary conditions compared with soft constraints.

### 3.3.3 Predicting the future responses with different numbers of observations

In this section, we compare the performance of theory-guided HCP and TgNN at different numbers of observations. Similar to the previous sections, the boxplot is utilized to evaluate each model. Figure 15 shows that, although the performance of different models has improved with an increase in the number of observations, the prediction accuracy and robustness of the theory-guided HCP are always higher than that of TgNN, irrespective of the number of observations.

In order to intuitively compare the performance of different models, the residual map of the predictions at the 30[th] time step of different models is provided in Figure 16. The abscissa and ordinate in the figure are the x and y coordinates of the flow field, respectively. Since only the first 18 time steps have observations, the predictions in Figure 16 are the results of extrapolation. The total number of observation points in the experiment are 0, 36, 180, and 1800, which means that the sample size of each time step is 0, 2, 10, and 100 since the same number of observations are sampled in each time step. In addition, the ANN is also considered in the experiment since the performance of ANN will be affected by the amount of observation data.

By comparing Figure 16, it can be seen that the performance of theory-guided HCP and TgNN is significantly better than that of the purely data-driven ANN model, which reflects that the theory-guided framework can effectively fuse information obtained from data and domain knowledge for model inference. Moreover, as the number of observations increases, the performance of all models has improved, but the residual of theory-guided HCP is always lower than that of TgNN, which indicates that hard constraints have stronger information extraction capabilities. It should be mentioned that as the number of observations increases, the accuracy of the ANN first increases and then decreases. This is mainly because the observation data come from the first 18 time steps and the predictions range from the $1^{st}$ time step to the $50^{th}$ time step. Therefore, the purely data-driven algorithm may overfit the data distribution of the first 18 time steps as the number of observations increase, which will affect the final overall prediction accuracy.

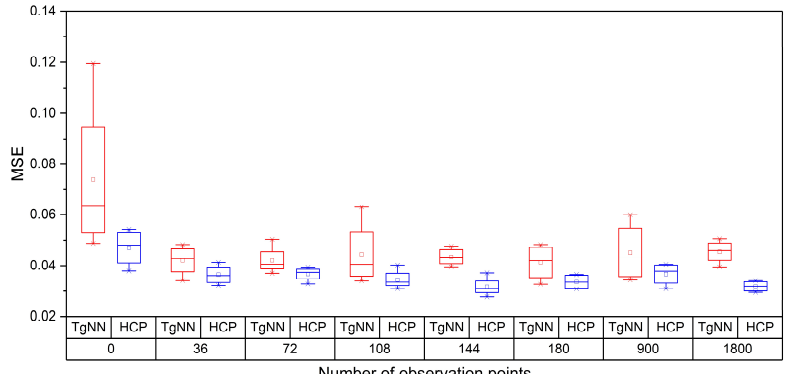

**Fig. 15.** Boxplot of the relative L2 loss of TgNN and HCP with different numbers of observation points.

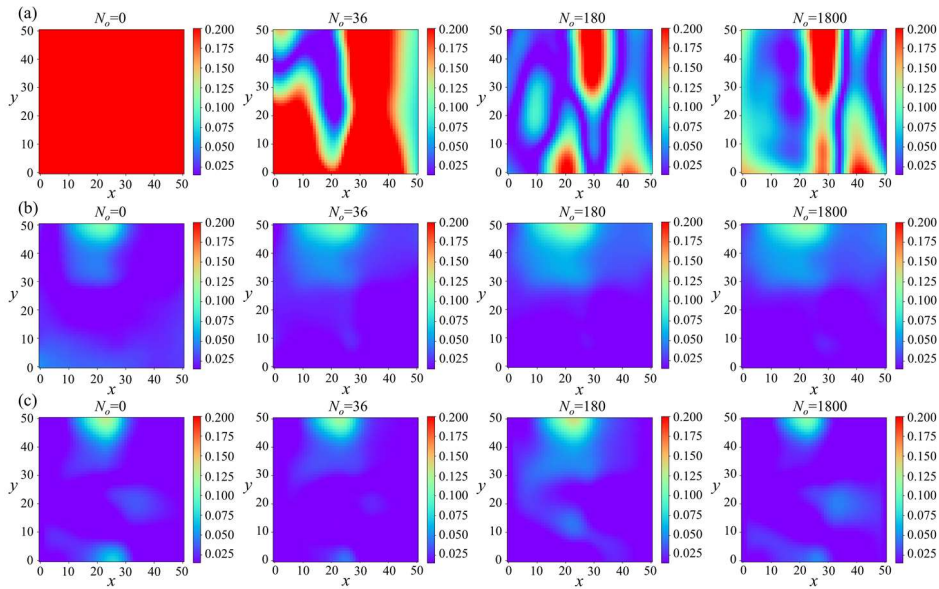

**Fig. 16.** Residual map of the predictions at the $30^{th}$ time step of (a) ANN; (b)TgNN; and (c) HCP.

## 3.4 Model robustness comparison
### 3.4.1 Predicting the future response in the presence of data noise

This section compares the performance of different models when noise exists in the observations. The noise conforms to a normal distribution with a mean value of 0 and a standard deviation of x% of the original undisturbed value. x% is called the noise level, which represents the magnitude of the disturbance. Figure 17 shows the effect of different noise levels on the observations. The noise gradually increases from Figure 17a (10%) to Figure 17d (60%).

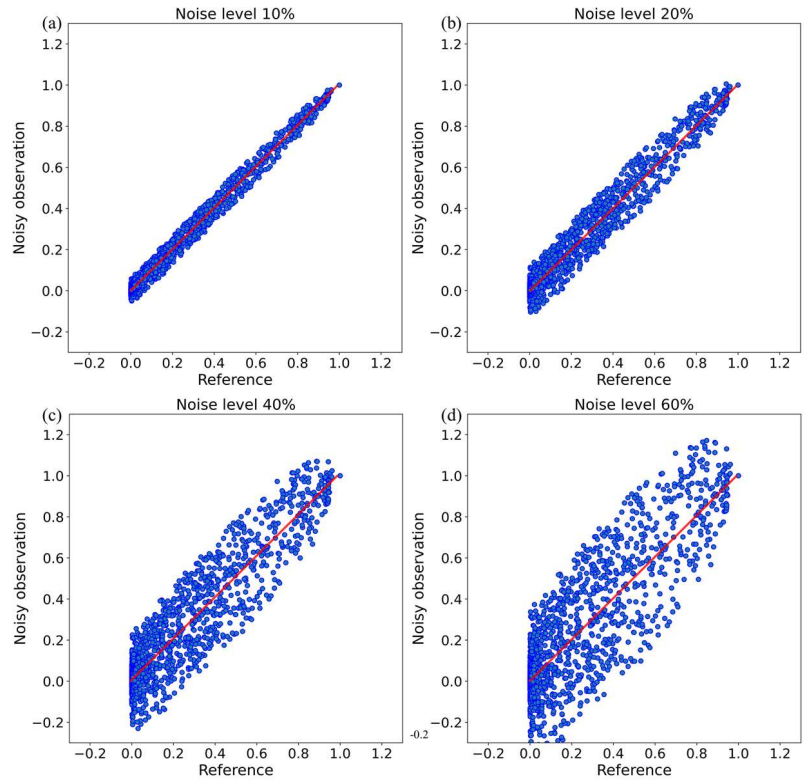

**Fig. 17.** Cross plot of observations with different noise levels: (a) 10% noise; (b) 20% noise; (c) 40% noise; and (d) 60% noise.

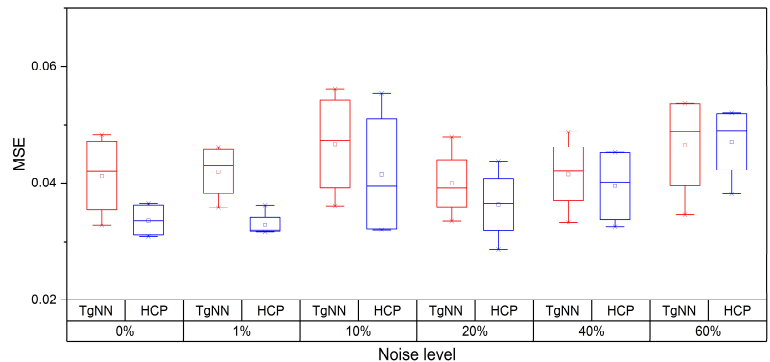

**Fig. 18.** Boxplot of the relative L2 loss of TgNN and HCP with different noise levels.

The performance of TgNN and theory-guided HCP under different noise levels is presented in Figure 18. When the noise level is less than 1%, theory-guided HCP has obvious higher accuracy and robustness. When the noise level further increases, however, the performance of theory-guided

HCP and TgNN will gradually converge. In the process of increasing the noise level from 10% to 60%, the model performance does not deteriorate significantly, which indicates that theory-guided HCP and TgNN are not sensitive to the noise level. Theoretically, the observations in the real world are composed of noise and real values. Purely data-driven models (e.g., ANN) cannot distinguish real values from noise, so they will be greatly disturbed. However, since the true value itself conforms to the physical mechanism, the introduction of the domain knowledge assists the TgNN and theory-guided HCP to distinguish noise in the data, which can effectively improve the model's robustness to noise.

**3.4.2 Predicting the future response in the presence of outliers**

This section compares the performance of different models with outliers. It should be emphasized that outliers are different from noise. Outliers are often caused by equipment failure or recording errors, while noise is common in observations and cannot be avoided. In other words, an outlier is an individual observation with gross error, while noise is only a perturbation that exists generally in the observations. Although the proportion of outliers in the sample is smaller, it tends to have a greater influence than noise. The value of outliers used in this experiment completely deviates from the value range of the observations. The outliers are randomly sampled from the interval [1,2], while the observations are normalized to [0,1] in the experiment. In order to clearly show the influence of outliers on the observations, Figure 19 presents the cross plot between the outliers and the true values at different proportions of outlier. It can be seen that the influence of outliers on data distribution is much greater than noise (Figure 17).

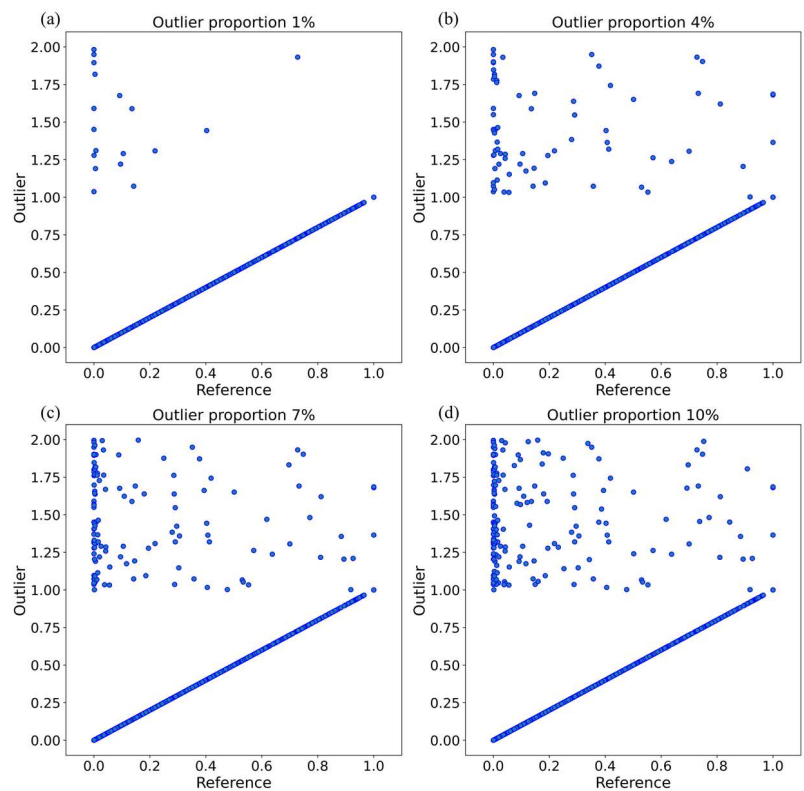

**Fig. 19.** Cross plot of observations with different proportion of outliers: (a) 1% outliers; (b) 4% outliers; (c) 7% outliers; and (d) 10% outliers.

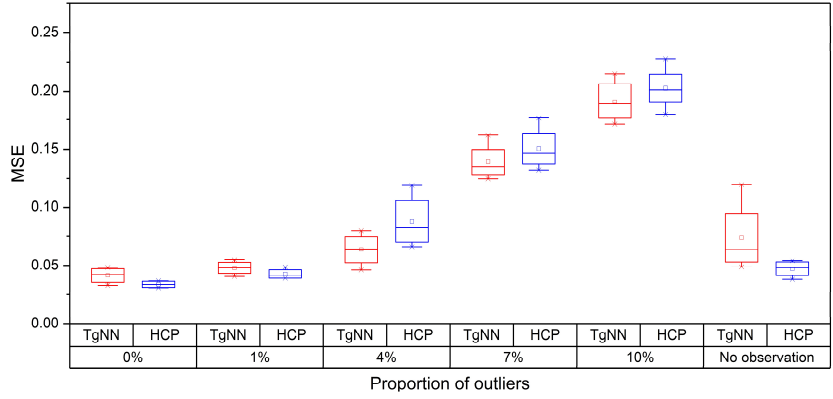

**Fig. 20.** Boxplot of the relative L2 loss of TgNN and HCP with different proportion of outliers.

The performance comparison of theory-guided HCP and TgNN under different proportions of outliers is shown in Figure 20. It can be seen that the prediction accuracy of both theory-guided HCP and TgNN decreases significantly as the number of outliers increases. When the proportion of outliers exceeds 4%, the performance of TgNN is slightly better than that of theory-guided HCP. Theoretically, when an outlier exists in the constraint patch, because the value of the outlier is markedly different from the other true values, it will cause significant deviation of the prediction matrix and projected prediction matrix in the constraint patch. The theory-guided HCP will adjust the data points in the constraint patch as a whole to satisfy the governing equation, and thus, the outlier will force the other points in the constraint patch to be projected as wrong values. Therefore, one outlier may destroy the entire constraint patch. Although the theory-guided HCP is sensitive to outliers, its performance is still better than conventional ANN (Appendix D shows that the error of theory-guided HCP is one order of magnitude smaller than that of ANN). Besides, outlier detection methods could be applied to data preprocessing before the projection to alleviate the influence of outliers on prediction accuracy and improve the model's robustness to outliers.

## 4. Conclusion

A theory-guided hard constraint projection (HCP) is proposed to introduce domain knowledge and prior information as hard constraints into neural networks through projection. Compared with commonly used methods, such as PINN and TgNN, that introduce domain knowledge as soft constraints through regularization terms of the loss function, the theory-guided HCP can ensure that the predicted values within the constraint patch strictly conform to the differential form of the governing equation (i.e., hard constraint). In principle, the projection based on hard constraints can not only describe the degree of deviation of the predictions from the governing equations at each iteration step, but also provide a strategy for adjusting the predictions to meet the physical constraints. Therefore, hard constraints are theoretically both more instructive and more efficient, which indicates higher information extraction efficiency. The major contributions include:

- The proposed theory-guided HCP is constructed under the theory-guided framework, which offers the advantage of being compatible with various forms of domain knowledge and prior information. Besides, a three-step method to embed hard constraints is proposed, which includes equation discretization, matrix decomposition, and projection. The theory-guided

HCP can use sparse observation, for example, based on the observations from early time steps in the flow field, it can predict the flow field at subsequent time steps under the indirect supervision of domain knowledge.

- Based on rigorous mathematical proofs, a general projection matrix $P = (I - A^T(AA^T)^{-1}A)$ is constructed based on constraint matrix $A$. This projection method can convert the prediction matrix on the constraint patch into the points on the hyperplane that satisfy the physical constraints. The HCP can be regarded as a non-trainable activation function determined by the governing equation in the neural network. In addition, theory-guided HCP introduces ghost cells to ensure that the projected results meet the boundary conditions.
- The experiments demonstrate that the method of introducing hard constraints based on projection can effectively reduce the algorithm's data demand and can achieve a higher prediction accuracy with fewer collocation points, boundary points, and observations. In the model robustness analysis experiment, it is confirmed that the theory-guided HCP is not sensitive to noise and can obtain stable prediction results under noise with a standard deviation of 40%.

The theory-guided HCP constitutes a preliminary attempt to integrate hard constraints and deep learning methods. Although the theory-guided HCP improves the learning ability of the model and can effectively use domain knowledge and prior information, it still possesses a limitation that require further investigation in the future. The current projection method can only guarantee that the prediction results in the constraint patch around the collocation point strictly conform to the physical constraints, but it cannot guarantee that the entire flow field is globally satisfied. In other words, theory-guided HCP is a local to global training method, but local compliance with constraints is a necessary and insufficient condition for global satisfaction. Only when the governing equation is met at all positions (constraint patches) of the flow field and the boundary and initial conditions are satisfied at the same time, can the overall physical constraints be guaranteed. Meanwhile, expanding the coverage of the constraint patch will lead to greater calculation time. When a single constraint patch is expanded to the entire physical field, it is equivalent to directly solving the governing equation. Therefore, how to balance the coverage of each constraint patch with computational efficiency is worthy of further study.

**Acknowledgements**
This work is partially funded by the National Natural Science Foundation of China (Grant no. 51520105005) and ROIS NII Open Collaborative Research 2021(20FC05).

**Appendix A: The proof of projection matrix**
**Theorem 1.** For $A \in \mathbb{R}^{m,n}$, where $m > n$, and $rank(A) = m$. The column space of $A$ is $C(A)$. The projection matrix of $C(A)$ is defined as $proj_{C(A)}$. Moreover, we have $proj_{C(A)} = A(A^TA)^{-1}A^T$.

**Proof.** The column space of $A$ is defined as $C(A)$ or $Range(A)$. For $Y^* \in C(A)$, we have: $A\theta = Y^*$.

The matrix $A$ is strictly skinny ($m > n$), $A\theta = Y^*$ is overdetermined, and it is difficult to find a set of $\theta^*$ that make the model fit all of the data points. Therefore, if we define the residual as $r = A\theta - Y$, we have $r \neq 0$.

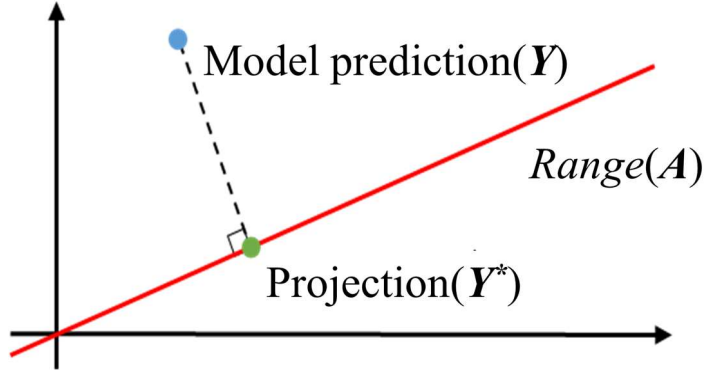

**Fig. A1.** Geometric interpretation of the projection on $Range(A)$.

We attempt to find a set of $\theta^*$ that minimizes the non-zero residual $r$. Minimizing the residual is equivalent to finding the projection of $Y$ on $C(A)$ or $Range(A)$, i.e., the point where $Y^*$ is closest to $Y$ on $Range(A)$, as shown in Figure A1. To minimize the residual, we calculate the gradient of the squared residual w.r.t. $\theta$, which is shown in Eq. A1:

$$\nabla_\theta r^2 = \nabla_\theta(\theta^T A^T A\theta - 2Y^T A\theta + Y^T Y) = 2A^T A\theta - 2AY \tag{A1}$$

It is known that when the residual is at its minimum, the gradient of the residual must be zero. Since the matrix $A^T A$ is usually invertible in practice, the $\theta^*$ can be obtained as follows:

$$\theta^* = (A^T A)^{-1} A^T Y \tag{A2}$$

The relationship between a given prediction value $Y$ and its projection $Y^*$ on $Range(A)$ can be constructed, as shown in Eq. A3:

$$Y^* = A\theta^* = A(A^T A)^{-1} A^T Y = proj_{C(A)} Y \tag{A3}$$

Therefore, for $A \in \mathbb{R}^{m,n}$, where $m > n$, and $rank(A) = m$. We have $proj_{C(A)} = A(A^T A)^{-1} A^T$. $\square$

**Theorem 2.** For $A \in \mathbb{R}^{m,n}$, where $m < n$, and $rank(A) = m$. Given any $w \in \mathbb{R}^n$, the closest point $w^*$ subject to $Aw = b$ is $w^* = (I - A^T(AA^T)^{-1}A)w + A^T(AA^T)^{-1}b$.

**Proof.** Suppose a subspace $S = \{\theta | A\theta = b\}$. For any $w \in \mathbb{R}^n$, we have $w = proj_S(w) + proj_{S^\perp}(w)$,

Since $S$ is the null space of $A$ and $S^\perp$ is the orthogonal subspace of $S$, we have the following:

$$S^\perp = N(A)^\perp = C(A^T) \tag{A4}$$

where $N(A)$ denotes the null space of A; and $C(A^T)$ is the columns space of $A^T$.

$A^T$ is a strictly skinny matrix and we have the conclusion that $proj_{C(A)} = A(A^T A)^{-1} A^T$ from Theorem 1. Therefore, $proj_{S^\perp}$ can be obtained as Eq. A5:

$$proj_{S^\perp} = proj_{C(A^\perp)} = A^T (AA^T)^{-1} A \tag{A5}$$

Substituting Eq. A5 into $w = proj_S(w) + proj_{S^\perp}(w)$, we can obtain Eq. A6:

$$proj_S = I - A^T (AA^T)^{-1} A \tag{A6}$$

Since $rank(A) = m$, we have $rank(AA^T) = m$, which is full rank. Therefore, $AA^T$ is invertible. Since $w^*$ is the projection of $w$ onto the subspace $S$, we have Eq. A7 according to the definition of projection:

$$proj_S(w - w^*) = (I - A^T (AA^T)^{-1} A)(w - w^*) = 0 \tag{A7}$$

Then, we can rewrite Eq. A6 as Eq. A8:

$$\begin{aligned} w^* &= (I - A^T (AA^T)^{-1} A)(w - w^*) + w^* \\ &= (I - A^T (AA^T)^{-1} A)w - (w^* - A^T (AA^T)^{-1} Aw^*) + w^* \\ &= (I - A^T (AA^T)^{-1} A)w + A^T (AA^T)^{-1} Aw^* \\ &= (I - A^T (AA^T)^{-1} A)w + A^T (AA^T)^{-1} b \end{aligned} \tag{A8}$$

Therefore, for $A \in \mathbb{R}^{m,n}$, where $m < n$, and $rank(A) = m$. Given any $w \in \mathbb{R}^n$, the closest point $w^*$ subject to $Aw = b$ is $w^* = (I - A^T(AA^T)^{-1}A)w + A^T(AA^T)^{-1}b$. □

It should be noted that, in Theorem 2, $Aw = b$ is not overdetermined, since $m < n$, which is different from Theorem 1 where $m > n$ in $A$. The problem in Theorem 1 can be regarded as solving an overdetermined matrix through the least squares method, where $m$ in $A \in \mathbb{R}^{m,n}$ is the number of constraints and it is equal to the batch size. However, in Theorem 2, $Aw = b$ only describes a single constraint patch, i.e., $m = 1$ and $m < n$ according to Eq. 9. Theoretically, because the number of constraints is less than the number of variables, $Aw = b$ has an infinite number of solutions in Theorem 2. Since $w$ denotes the neural network predictions (i.e., $H$ in Eq. 8), however, there is no guarantee that $AH = 0$ or $Aw = b$ holds. It is still necessary to find the projection of $w$ on the constrained hyperplane.

In order to improve the computational efficiency, the mini-batch method is used to train the theory-guided HCP. It is incorrect to set $m$ (the number of constraints) to the batch size when using mini-batch since there is only one constraint in the constraint patch. Instead, we should transform the matrix $A \in \mathbb{R}^{m,n}$ in Theorem 2 into a three-dimensional matrix $A \in \mathbb{R}^{b,m,n}$, where $b$ denotes the batch size.

**Appendix B: Projection strategy based on ghost cell to process boundary conditions**

This study focuses on the commonly-used constant pressure boundary and no-flow boundary. The projections of other boundary conditions can also be achieved using a similar method. First, we introduce the concept of the ghost cell in computational fluid dynamics (CFD) to deal with the boundary conditions. Ghost cells are virtual grid cells outside of the boundary, similar to the padding method in computer vision, as shown in Figure B1. By specifying the relationship between the ghost cells and the boundary cells, we can constrain the boundary cells to obtain a solution that satisfies the boundary conditions.

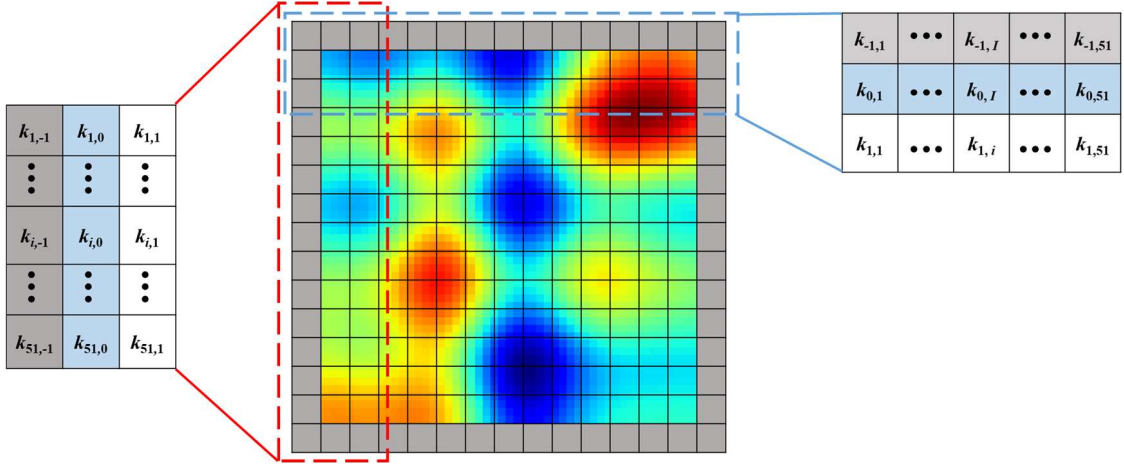

**Fig. B1.** Illustration of ghost cells (grey) and boundary cells (blue).

For the constant pressure boundary, it can be ensured that the projected predictions not only satisfy the governing equation in the constraint patch, but also have a given pressure on the boundary by adjusting the values of hydraulic pressure and hydraulic conductivity at the ghost cells. For example, if the left boundary of a flow field has a constant pressure boundary (Figure B1), then the boundary cells (blue), the ghost cells (grey), and the cells on the right (white) need to obey a specific pressure and hydraulic conductivity relationship. In other words, the pressure and hydraulic conductivity in the ghost cells (grey) are determined by the boundary cells (blue) and the cells on the right side of the boundary (white). In addition, the flow rate at the constant pressure boundary is perpendicular to the isopleth of pressure. Therefore, the flow rate on the left and right sides of the boundary should be equal, according to the conservation of mass. In order to ensure that the flow field has a constant pressure boundary, the pressure (hydraulic head) and hydraulic conductivity at the ghost cells should conform to the relationship described by Eq. B1:

$$\begin{aligned} k_{i,-1} &= k_{i,1} \\ h_{i,-1} &= 2h_{i,0} - h_{i,1} \end{aligned} \tag{B1}$$

For the no-flow boundary, the upper boundary of the flow field is taken as an example, as shown in Figure B1. The no-flow boundary is easier to handle than the constant pressure boundary, as the no-flow boundary can be guaranteed by simply setting the hydraulic conductivity at the ghost cells to 0, which is shown in Eq. B2:

$$k_{i,-1} = 0 \tag{B2}$$

Therefore, the constraint matrix and prediction matrix at the boundary of a flow field can be modified prior to the projection by Eq. B1 and Eq. B2. Subsequently, hard constraint projection is performed to obtain physically reasonable predictions that meet the boundary conditions and satisfy the governing equation simultaneously.

**Appendix C: Cost of theory-guided HCP and simulation to solve a physical field**

The comparative experiment of computational cost of simulation and neural network is shown in Figure C1. It can be seen that as the time step increases, the computational cost of the simulation increases linearly, while the computational cost of the neural network remains basically unchanged. This is because the simulation needs to calculate from the 1$^{st}$ time step to the $t^{th}$ time step and the neural network only needs to input the coordinate (t, x, y) as independent variables into the model to obtain the results. Theoretically speaking, differing from the numerical simulation, the training and inference process of neural network are separated. Although the calculation time of the training process is relatively high, only the low time-consuming inference process is required in actual use since the model is pre-trained in practice (e.g., the inference time based on GTX1080Ti is 0.028 seconds for the case of 5000 time steps). Therefore, the neural networks are especially suitable for practical applications. In fact, when predicting the data at the 50$^{th}$ time step, the simulation time is 32 times that of the neural network, and for the 5000$^{th}$ time step, the simulation time is 445 times that of the neural network in the experiment.

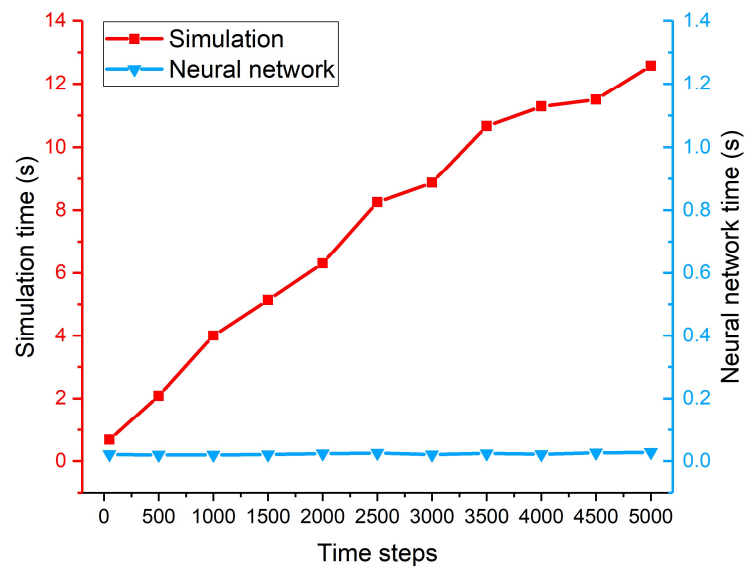

**Fig. C1.** Computational time of simulation (red and left axis) and neural network (blue and right axis).

# Appendix D: Experiment results of ANN, TgNN, and HCP

**Table D1.** Relative L2 loss of ANN, TgNN, and theory-guided HCP in different scenarios.

| | | ANN | TgNN | HCP |
|---|---|---|---|---|
| Number of collocation points | 40 | 0.213±0.083 | 0.275±0.121 | **0.184**±0.144 |
| | 70 | 0.213±0.083 | 0.138±0.081 | **0.091**±0.027 |
| | 90 | 0.213±0.083 | 0.083±0.024 | **0.072**±0.005 |
| | 100 | 0.213±0.083 | 0.081±0.017 | **0.062**±0.007 |
| | 180 | 0.213±0.083 | 0.059±0.019 | **0.042**±0.008 |
| | 400 | 0.213±0.083 | 0.051±0.010 | **0.035**±0.006 |
| | 700 | 0.213±0.083 | 0.043±0.004 | **0.035**±0.002 |
| | 1000 | 0.213±0.083 | 0.041±0.006 | **0.034**±0.003 |
| | 1800 | 0.213±0.083 | 0.049±0.011 | **0.032**±0.003 |
| Number of boundary points | 1 | 0.213±0.083 | 0.183±0.039 | **0.117**±0.024 |
| | 4 | 0.213±0.083 | 0.084±0.019 | **0.041**±0.006 |
| | 7 | 0.213±0.083 | 0.061±0.010 | **0.034**±0.004 |
| | 10 | 0.213±0.083 | 0.056±0.005 | **0.035**±0.005 |
| | 100 | 0.213±0.083 | 0.041±0.005 | **0.034**±0.004 |
| | 400 | 0.213±0.083 | 0.045±0.005 | **0.032**±0.005 |
| | 700 | 0.213±0.083 | 0.040±0.004 | **0.031**±0.003 |
| | 1000 | 0.213±0.083 | 0.038±0.003 | **0.035**±0.007 |
| | 10000 | 0.213±0.083 | 0.041±0.006 | **0.034**±0.003 |
| Number of observation points | 0 | 4.017±1.710 | 0.074±0.027 | **0.047**±0.006 |
| | 32 | 0.406±0.176 | 0.042±0.005 | **0.036**±0.003 |
| | 76 | 0.234±0.034 | 0.042±0.005 | **0.037**±0.002 |
| | 108 | 0.202±0.062 | 0.045±0.011 | **0.035**±0.003 |
| | 144 | 0.205±0.079 | 0.044±0.003 | **0.032**±0.003 |
| | 180 | 0.213±0.083 | 0.041±0.006 | **0.034**±0.003 |
| | 900 | 0.239±0.054 | 0.045±0.010 | **0.037**±0.004 |
| | 1800 | 0.245±0.073 | 0.046±0.004 | **0.032**±0.002 |
| Noise level | 1% | 0.207±0.077 | 0.042±0.004 | **0.033**±0.002 |
| | 10% | 0.272±0.051 | 0.047±0.008 | **0.042**±0.010 |
| | 20% | 0.339±0.085 | 0.040±0.005 | **0.036**±0.005 |
| | 40% | 0.896±0.489 | 0.042±0.006 | **0.040**±0.006 |
| | 60% | 1.838±0.871 | **0.047**±0.008 | **0.047**±0.006 |
| Proportion of outliers | 1% | 1.114±0.357 | 0.048±0.005 | **0.043**±0.004 |
| | 4% | 3.621±2.372 | **0.064**±0.013 | 0.088±0.020 |
| | 7% | 2.455±0.929 | **0.139**±0.014 | 0.150±0.016 |
| | 10% | 1.645±0.597 | **0.191**±0.017 | 0.202±0.017 |

**References**

1. Zhang, B., Zhu, J., and Su, H., *Towards the third generation artificial intelligence.* SCIENTIA SINICA Informationis, 2020. **50**(09): p. 1281-1302.
2. Laporte, F., *On the design of an expert system guide for the use of scientific software.* Computer methods in applied mechanics engineering, 1989. **75**(1-3): p. 241-250.

3. Liao, S.-H., *Expert system methodologies and applications—a decade review from 1995 to 2004.* Expert Systems with Applications, 2005. **28**(1): p. 93-103.
4. Studer, R., Benjamins, V.R., and Fensel, D., *Knowledge engineering: principles and methods.* Data Knowledge Engineering, 1998. **25**(1-2): p. 161-197.
5. Lederberg, J., Feigenbaum, E.A., Buchanan, B.G., and Lindsay, R.K., *Applications of Artificial Intelligence for Organic Chemistry: The DENDRAL Project.* 1980, New York: McGraw-Hill Companies.
6. Buchanan, B.G. and Shortliffe, E.H., *Rule- Based Expert Systems : The MYCIN Experiments of the Stanford Heuristic Programming Project.* 1984, Boston: Addison-Wesley.
7. Campbell, M., Hoane Jr, A.J., and Hsu, F.-h., *Deep blue.* Artificial intelligence, 2002. **134**(1-2): p. 57-83.
8. Rosenblatt, F., *The perceptron: a probabilistic model for information storage and organization in the brain.* Psychological Review, 1958. **65**(6): p. 386.
9. Fukushima, K., *A self-organizing neural network model for a mechanism of pattern recognition unaffected by shift in position.* Biol Cybernet, 1980. **36**: p. 193-202.
10. Hopfield, J., *Neural networks and physical systems with emergent collective computational abilities.* Proceedings of the National Academy of Sciences, 1982. **79**(8): p. 2554-2558.
11. Hochreiter, S. and Schmidhuber, J., *Long short-term memory.* Neural Computation, 1997. **9**(8): p. 1735-1780.
12. Kipf, T.N. and Welling, M., *Semi-supervised classification with graph convolutional networks.* arXiv preprint arXiv:.02907, 2016.
13. Samaniego, E., Anitescu, C., Goswami, S., Nguyen-Thanh, V.M., Guo, H., Hamdia, K., et al., *An energy approach to the solution of partial differential equations in computational mechanics via machine learning: Concepts, implementation and applications.* Computer Methods in Applied Mechanics Engineering, 2020. **362**: p. 112790.
14. Sheikholeslami, M., Gerdroodbary, M.B., Moradi, R., Shafee, A., and Li, Z., *Application of Neural Network for estimation of heat transfer treatment of Al2O3-H2O nanofluid through a channel.* Computer Methods in Applied Mechanics Engineering, 2019. **344**: p. 1-12.
15. Dong, Y., Su, H., Zhu, J., and Bao, F., *Towards interpretable deep neural networks by leveraging adversarial examples.* arXiv preprint arXiv:.05493, 2017.
16. Wang, N., Zhang, D., Chang, H., and Li, H., *Deep learning of subsurface flow via theory-guided neural network.* Journal of Hydrology, 2020. **584**: p. 124700.
17. Baker, N., Alexander, F., Bremer, T., Hagberg, A., Kevrekidis, Y., Najm, H., et al., *Workshop report on basic research needs for scientific machine learning: Core technologies for artificial intelligence.* 2019, USDOE Office of Science (SC), Washington, DC (United States).
18. Xia, Y., Leung, H., and Wang, J., *A projection neural network and its application to constrained optimization problems.* IEEE Transactions on Circuits Systems I: Fundamental Theory Applications, 2002. **49**(4): p. 447-458.
19. Gao, X.-B., Liao, L.-Z., and Qi, L., *A novel neural network for variational inequalities with linear and nonlinear constraints.* IEEE Transactions on Neural Networks, 2005. **16**(6): p. 1305-1317.
20. Pathak, D., Krahenbuhl, P., and Darrell, T. *Constrained convolutional neural networks for weakly supervised segmentation.* in *Proceedings of the IEEE international conference on computer vision.* 2015.

21. Raissi, M., Perdikaris, P., and Karniadakis, G., *Physics-informed neural networks: A deep learning framework for solving forward and inverse problems involving nonlinear partial differential equations.* Journal of Computational Physics, 2019. **378**: p. 686-707.
22. Wessels, H., Weißenfels, C., and Wriggers, P., *The neural particle method–An updated Lagrangian physics informed neural network for computational fluid dynamics.* Computer Methods in Applied Mechanics Engineering, 2020. **368**: p. 113127.
23. Sit, M., Demiray, B.Z., Xiang, Z., Ewing, G.J., Sermet, Y., and Demir, I., *A comprehensive review of deep learning applications in hydrology and water resources.* Water Science Technology, 2020.
24. He, T. and Zhang, D., *Deep learning of dynamic subsurface flow via theory-guided generative adversarial network.* arXiv preprint arXiv:2006.13305, 2020.
25. Karpatne, A., Atluri, G., Faghmous, J.H., Steinbach, M., Banerjee, A., Ganguly, A., et al., *Theory-guided data science: A new paradigm for scientific discovery from data.* IEEE Transactions on Knowledge and Data Engineering, 2017. **29**(10): p. 2318-2331.
26. Karpatne, A., Watkins, W., Read, J., and Kumar, V., *Physics-guided neural networks (pgnn): An application in lake temperature modeling.* arXiv preprint arXiv:.11431, 2017.
27. Beucler, T., Pritchard, M., Rasp, S., Ott, J., Baldi, P., and Gentine, P., *Enforcing analytic constraints in neural-networks emulating physical systems.* arXiv preprint arXiv:.00912, 2019.
28. Chen, Y. and Zhang, D., *Theory guided deep-learning for load forecasting (TgDLF) via ensemble long short-term memory.* EnerarXiv, 2020.
29. Xing, L., Dongxiao, Z., and and Xu, Z., *Deep Learning Based Forecasting of Photovoltaic Power Generation via Theory-guided LSTM.* EnerarXiv, 2020.
30. Xu, K. and Darve, E., *Physics constrained learning for data-driven inverse modeling from sparse observations.* arXiv preprint arXiv:.10521, 2020.
31. Mohan, A.T., Lubbers, N., Livescu, D., and Chertkov, M., *Embedding hard physical constraints in neural network coarse-graining of 3d turbulence.* arXiv preprint arXiv:.00021, 2020.
32. Long, Z., Lu, Y., Ma, X., and Dong, B. *Pde-net: Learning pdes from data*. in *International Conference on Machine Learning*. 2018. PMLR.
33. Chen, Y. and Zhang, D., *Physics-constrained indirect supervised learning.* arXiv preprint arXiv:.14293, 2020.
34. Fukunaga, K. and Koontz, W.L., *Application of the Karhunen-Loeve expansion to feature selection and ordering.* IEEE Transactions on computers, 1970. **100**(4): p. 311-318.
35. Phoon, K., Huang, S., and Quek, S., *Simulation of second-order processes using Karhunen–Loeve expansion.* Computers Structures, 2002. **80**(12): p. 1049-1060.
36. Harbaugh, A.W., *MODFLOW-2005, the US Geological Survey Modular Ground-Water Model: the Ground-Water Flow Process*. 2005: US Department of the Interior, US Geological Survey Reston, VA.